\documentclass[sigconf]{acmart}
\usepackage{subfigure}
\usepackage{amsmath}
\usepackage{bm}
\usepackage[noend]{algpseudocode}
\usepackage{algorithm}
\algrenewcommand{\algorithmicrequire}{\textbf{Input:}}
\algrenewcommand{\algorithmicensure}{\textbf{Output:}}
\algnewcommand\algorithmicforeach{\textbf{for each}}
\algdef{S}[FOR]{ForEach}[1]{\algorithmicforeach\ #1\ \algorithmicdo}
\usepackage{arydshln}
\usepackage{booktabs}
\usepackage{multirow}
\usepackage{marvosym}
\usepackage{ifsym}
\usepackage{stfloats}

\AtBeginDocument{%
  \providecommand\BibTeX{{%
    \normalfont B\kern-0.5em{\scshape i\kern-0.25em b}\kern-0.8em\TeX}}}

\setcopyright{acmcopyright}
\copyrightyear{2022} 
\acmYear{2022} 
\setcopyright{acmlicensed}\acmConference[MM '22]{Proceedings of the 30th ACM International Conference on Multimedia}{October 10--14, 2022}{Lisboa, Portugal}
\acmBooktitle{Proceedings of the 30th ACM International Conference on Multimedia (MM '22), October 10--14, 2022, Lisboa, Portugal}
\acmPrice{15.00}
\acmDOI{10.1145/3503161.3548284}
\acmISBN{978-1-4503-9203-7/22/10}

\acmSubmissionID{2305}

\begin{document}

\title{Chunk-aware Alignment and Lexical Constraint for Visual Entailment with Natural Language Explanations}






\author{Qian Yang}
\authornote{Both authors contributed equally to this research.}
\email{yangqianhitsz@163.com}
\affiliation{%
\institution{Harbin Institute of Technology}
  \city{Shenzhen}
  \country{China}
}

\author{Yunxin Li}
\authornotemark[1]
\email{liyunxin987@163.com}
\affiliation{%
\institution{Harbin Institute of Technology}
  \city{Shenzhen}
  \country{China}
}

\author{Baotian Hu}
\email{hubaotian@hit.edu.cn}
\authornote{Corresponding author.}
\affiliation{%
\institution{Harbin Institute of Technology}
  \city{Shenzhen}
  \country{China}
}

\author{Lin Ma}
\authornotemark[2]
  \email{forest.linma@gmail.com}
\affiliation{%
\institution{Meituan}
  \city{Beijing}
  \country{China}
}

\author{Yuxin Ding}
  \email{yxding@hit.edu.cn}
\affiliation{%
\institution{Harbin Institute of Technology}
  \city{Shenzhen}
  \country{China}
}

\author{Min Zhang}
\email{zhangmin2021@hit.edu.cn}
\affiliation{%
\institution{Harbin Institute of Technology}
  \city{Shenzhen}
  \country{China}
}

\renewcommand{\shortauthors}{Qian Yang et al.}

\begin{abstract}

Visual Entailment with natural language explanations aims to infer the relationship between a text-image pair and generate a sentence to explain the decision-making process. Previous methods rely mainly on a pre-trained vision-language model to perform the relation inference and a language model to generate the corresponding explanation. However, the pre-trained vision-language models mainly build token-level alignment between text and image yet ignore the high-level semantic alignment between the phrases (chunks) and visual contents, which is critical for vision-language reasoning.
Moreover, the explanation generator based only on the encoded joint representation does not explicitly consider the critical decision-making points of relation inference. Thus the generated explanations are less faithful to visual-language reasoning.
To mitigate these problems, we propose a unified \textbf{C}hunk-aware \textbf{A}lignment and \textbf{Le}xical \textbf{C}onstraint based method, dubbed as \textbf{CALeC}.
It contains a \textbf{C}hunk-aware \textbf{S}emantic \textbf{I}nteractor (arr. \textbf{CSI}), a relation inferrer, and a \textbf{Le}xical \textbf{C}onstraint-aware \textbf{G}enerator (arr. \textbf{LeCG}). Specifically, CSI exploits the sentence structure inherent in language and various image regions to build chunk-aware semantic alignment. Relation inferrer uses an attention-based reasoning network to incorporate the token-level and chunk-level vision-language representations. LeCG utilizes lexical constraints to expressly incorporate the words or chunks focused by the relation inferrer into explanation generation, improving the faithfulness and informativeness of the explanations. We conduct extensive experiments on three datasets, and experimental results indicate that CALeC significantly outperforms other competitor models on inference accuracy and quality of generated explanations. Code is available here:  \href{https://github.com/HITsz-TMG/ExplainableVisualEntailment}{https://github.com/HITsz-TMG/ExplainableVisualEntailment}.
\end{abstract}

\ccsdesc[500]{Computing methodologies~Computer vision tasks}
\ccsdesc[300]{Computing methodologies~Natural language generation}

\keywords{Visual Entailment; Explanation Generation; Semantic Alignment}

\maketitle
\newcommand{\fixme}[1]{\textcolor{red}{\textbf{yunxin: }\xspace#1}\xspace}
\section{Introduction}
\begin{figure}[t]
  \centering
  \includegraphics[width=\linewidth]{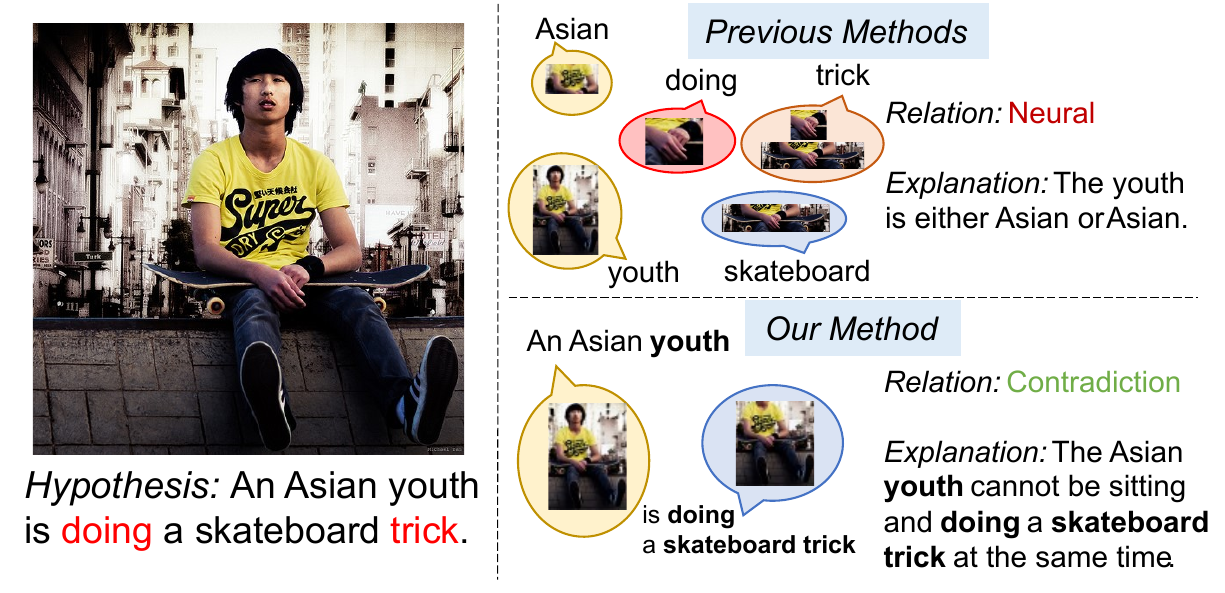}
  \caption{Visual comparison between our method and previous methods. Our method can accurately align ``is doing a skateboard trick'' to the semantically relevant region rather than separate objects like previous methods. It can also generate an explanation centered around the keywords of the inference process, which is more informative and faithful than previous methods.
}
  \label{fig:intro_case}
\end{figure}
Visual Entailment with Natural Language Explanations (VE-NLE) aims to infer the relationship (\textit{entailment, contradiction, neutral}) between a text-image pair and generate an explanation that can reflect the decision-making process.
Compared to conventional image-text matching tasks~\cite{flickr30k,mscoco}, Visual Entailment (VE) requires discerning more fine-grained cross-modal information on the input pair, because ``\textit{neutral}'' needs the model to conclude the uncertainty between ``\textit{entailment (yes)}'' and ``\textit{contradiction (no)}''.
Moreover, input text in VE contains more abundant information related to the image than Visual Question Answering~\cite{VQA2,coco_qa,visual7w}. Thus, it requires more precise sentence comprehension and proper visual grounding to infer the relationship.
Natural language explanations (NLE) could help correct the model bias and understand the decision-making process~\cite{evil,GPT_NLE} in a human-friendly way.
And a convincing explanation should center around the input text-image pair and reflect the inference process faithfully.

For VE-NLE, typical methods ~\cite{VQAX,Faithful_NLE,Vision_info_explanation,evil} adopt a vision-language model to obtain the inference result via learning a joint representation of the input pair. The representation is fed to a language model to generate the corresponding explanation.
Despite improvements on inference accuracy and explanation quality, these works still have certain limitations.
First, most vision-language models~\cite{MCB,chen2020uniter,oscar,soho} focus on building token-level alignment to learn the joint representation, neglecting the high-level semantic alignment between the phrase and image region. It often leads to ambiguous semantic understanding and vague alignment, making the classifier error-prone. For example, as previous methods shown in Figure~\ref{fig:intro_case}, although they align the words (e.g., youth, doing, skateboard) to different image regions, the discrete token-level alignments can not capture that the chunk describes an action ``\textit{is \textbf{doing} a skateboard \textbf{trick}}''. The misinterpretation of the chunk leads to the incorrect alignment of \textit{doing} and \textit{trick} and results in the wrong inference result.
Secondly, the above methods generate explanations by solely applying attention to the joint representation, neglecting the critical decision-making points of relation inference. Thus, the explanation is easily confined to limited words, e.g., the explanation only attends to ``\textit{Asian youth}'' and is irrelevant to the second half of the input  ( see the explanation generated by previous methods in Figure~\ref{fig:intro_case}).
To enhance the correlation between relation inference and generation, \citet{exlpanation_refine} and \citet{GPT_NLE} utilize a language model to generate the inference result and explanation in a sequence simultaneously. Nevertheless, they can only attend to the results in each step of explanation generation, and the vital information of the inference process is still neglected.

To tackle the two problems, we propose a unified \textbf{C}hunk-aware \textbf{A}lignment and \textbf{Le}xical \textbf{C}onstraint based method (\textbf{CALeC}).
CALeC contains a \textbf{C}hunk-aware \textbf{S}emantic \textbf{I}nteractor (\textbf{CSI}), a relation inferrer and a \textbf{Le}xical \textbf{C}onstraint-aware \textbf{G}enerator (\textbf{LeCG}).
CSI exploits the rich semantics contained in chunks. It adopts a within-chunk interactor and an inter-chunk interactor to learn chunk-level semantics. Then it utilizes a cross-modal interactor to build alignments between chunks and regions, removing the ambiguous alignments.
Relation inferrer combines the outputs of CSI and a pre-trained vision-language model to gain a comprehensive representation of the input text-image pair and utilizes an attention-based reasoning network to predict the relation.
After that, LeCG generates explanations related to the inference process and input.
It utilizes a transformer-based generator to obtain the initial generation probability conditioned on the encoded representation and inferred result. Then LeCG chooses the keywords with higher attention weight during inference as the lexical constraint set and gains a lexical constraint probability over this set. The two probabilities are combined to generate the explanation.
Moreover, we utilize a constrained beam sample during testing to score each beam with the probability and number of constraint words occurrences.

We conduct extensive experiments on the current biggest VE-NLE dataset e-SNLI-VE~\cite{evil}. To demonstrate the generalizability of CALeC to other vision-language tasks, we also report results on two VQA-NLE datasets, VQA-X~\cite{VQAX} and VCR~\cite{VCR}.
Experimental results show that CALeC surpasses the previous state-of-the-art method on a wide range of automatic evaluation metrics. Our quality analysis indicates that the generated explanations of CALeC improve on the aspects of faithfulness and relevance.

In summary, the contributions of our work are three-fold: 
\textit{1)} We propose a unified chunk-aware alignment and lexical constraint based method (CALeC), which contains a chunk-aware semantic interactor (CSI) to exploit the rich semantics of chunks, a relation inferrer to obtain relations, and a lexical constraint-aware generator (LeCG) to produce correlated explanations according to the inference process and input.
\textit{2)} We introduce CSI and LeCG. CSI explicitly utilizes the chunks and various image regions to build the chunk-aware semantic alignment. LeCG incorporates keywords related to inference results to generate faithful explanations.
\textit{3)} Experimental results show that CALeC remarkably surpasses existing methods for inference accuracy and explanation faithfulness on the VE-NLE dataset. It also generalizes well across two VQA-NLE datasets.

\section{Related Works}

To help decrease the class bias and enhance the ability of fine-grained reasoning, \citet{vte} build the visual entailment dataset SNLI-VE, which combines Stanford Natural Language Inference (SNLI)~\cite{snli} and Flickr30k~\cite{flickr30k}.
They design a two-stream attention network to model the fine-grained cross-modal reasoning and demonstrate their interpretability via attention visualizations.
To explain the decision-making process more human-friendly and detailed, \citet{evil} propose combining visual entailment with natural language explanations and building the first VE-NLE dataset e-SNLI-VE, which is also the current largest NLE dataset for vision-language tasks.
Based on it, they establish a benchmark e-ViL for vision-language tasks with NLE, which contains e-SNLI-VE and two VQA-NLE datasets: VQA-X \cite{VQAX} and VCR \cite{VCR}.

\textbf{Inference Accuracy.}
Some works focus on improving inference accuracy.
\citet{VQAX} combine multi-modal information via bilinear pooling to predict the answer and utilize an LSTM-based language model to generate the explanation conditioned on the pooling representations.
\citet{evil} propose e-UG that adopts a powerful pre-trained vision-language model UNITER \cite{chen2020uniter} to learn joint representations and GPT-2 \cite{GPT2} to generate explanations.
However, though existing pre-trained models~\cite{chen2020uniter,soho,simvlm,unified_mm_pretrain} have made progress in inference accuracy, the general sequential attentive models focus on building token-level alignment, neglecting the rich semantics contained in phrase.

\textbf{Explanation Faithfulness.}
Other works focus on getting more faithful explanations.
\citet{Faithful_NLE} filter out the samples whose visual explanation does not relevant to the predicted answer via GradCAM~\cite{grad_cam}. They utilize an improved Up-Down VQA model~\cite{vqa_up_bottom} for answer inference, and an Up-Down LSTM model~\cite{vqa_up_bottom} to generate explanations.
\citet{Vision_info_explanation} use different vision-language models to obtain expressive text-image representations, and feed the encoded representations to GPT-2~\cite{GPT2} to generate explanations. 
However, the language model can only attend to input pairs via attention, treating inference and generation as separate tasks.
To enhance the correlation between inference and explanation generation, \citet{robust_explanation} use a mutually collaborating module to conduct a jointly adversarial attack on answering and explanation, which also helps improve the robustness of the model. 
\citet{exlpanation_refine} and \citet{GPT_NLE} convert inference as a text generation task and utilize the language model to generate the answer and explanation simultaneously.
Nevertheless, they fail to incorporate the critical words of the inference into generation.

\begin{figure*}[ht]
  \centering
  \includegraphics[width=\linewidth]{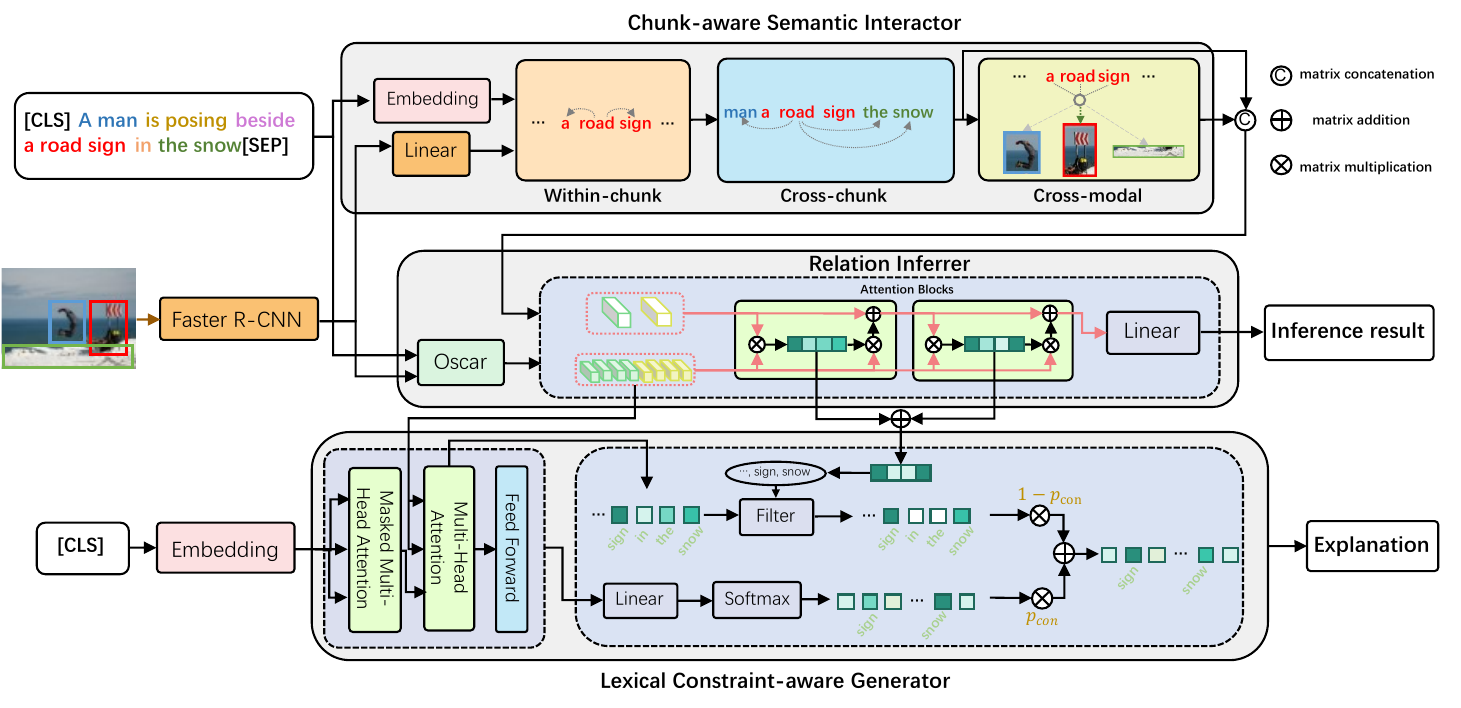}
  \caption{The overall architecture of CALeC for VE-NLE. Words of the input text in the same chunk are depicted using the same color. First, the chunk-aware semantic interactor exploits rich semantics contained in chunks to build chunk-aware semantic alignment. Then, the relation inferrer uses an attention-based reasoning network to incorporate the token-level and chunk-level vision-language representations. Finally, the lexical constraint-aware generator incorporates the keywords during inference into the explanation to improve the relevance and faithfulness.
  }
  \label{fig:model}
\end{figure*}
\section{Methodology}

Given an image premise $I$ and a text hypothesis $T$, VE-NLE aims to infer the relationship $y$ between the input pair and generate an explanation $E$ that can reflect the decision-making process.
An accurate inference requires a precise understanding of the sentence, and a convincing explanation needs to reflect the inference process faithfully.
To this end, we propose CALeC, whose architecture is shown in Figure~\ref{fig:model}.
It consists of three key components: a chunk-aware semantic interactor that exploits the rich semantics contained in phrases (Section~\ref{inference}); a relation inferrer that conducts fine-grained inference on the combined outputs of the chunk-aware semantics interactor and a pre-trained vision-language model (Section~\ref{Reasoning_module}); a lexical constraint-aware generator that incorporates the keywords during inference into explanation to improve the faithfulness (Section~\ref{generator}).
We train CALeC separately and utilize a constrained beam sample during testing. 
The training and testing detail descriptions are in Section~\ref{training}.

\subsection{Chunk-aware Semantic Interactor}
\label{inference}

We utilize a chunk-aware semantic interactor (CSI) to exploit semantics in phrases and build chunk-level alignment.
CSI takes the concatenation of text and image $[T;I]$ as the input.

For text $T=(w_1, w_2, ..., w_M)$, we add special tokens [CLS] and [SEP] to denote the start and end of the text.
The tokens are fed into an embedding layer to get $\textbf{H}^w=(\textbf{h}_{CLS}, \textbf{h}^w_{1}, ..., \textbf{h}^w_{M}, \textbf{h}_{SEP})$, where $\textbf{h}^w_{i}$ is the representation of $i$-th token.
For image $I$ with $N$ regions, we utilize a pre-trained Faster R-CNN~\cite{fasterRCNN} to extract the global image feature and region features.
We feed them into a fully-connected layer, obtaining the final representation $\textbf{H}^r=(\textbf{h}_g, \textbf{h}^r_{1}, ..., \textbf{h}^r_{N})$, where $\textbf{h}_g$ is the representation of the global image and $\textbf{h}^r_{i}$ is the representation of the $i$-th region.
We use a tagging model to get the borders $B$ of text chunks.
$B$ contains the start index and end index of each chunk, which can be formulated as $B=([b_{1,s},b_{1,e}], ..., [b_{K,s},b_{K,e}])$, $b_{i,s}$ and $b_{i,e}$ denotes the start index and end token index of $i$-th chunk, respectively.
Then, three interactors of different levels are used to learn the chunk-aware representations.

\subsubsection{Within-chunk Semantic Interactor}
We first adopt a within-chunk semantic interactor to exploit the rich semantics contained in chunks.
Each token has access only to the tokens in the same chunk group as itself.
For example, as showed in Figure~\ref{fig:model}, \textit{``road''} can only attend to \textit{``a road sign''}.
Specifically, for the $i$-th token $w_i$ in the $k$-th chunk, we calculate its attention score to the $j$-th token $w_j$ as follows,
\begin{equation}
    u_{ij} =\begin{cases}
    \frac{({\textbf{h}_{i}^w}{\textbf{W}_1^{q}})({\textbf{h}_j^w}{\textbf{W}_1^{k}})^\mathsf{T}}{\sqrt{d}}, &\text{where $j \in [b_{k,s},b_{k,e}]$}
    \cr -\infty, &\text{where $j \notin [b_{k,s},b_{k,e}]$}
    \end{cases}
\end{equation}
where $\textbf{h}_{i}^w$ denotes the representation of $i$-th token, $d$ denotes the dimension of the $\textbf{h}$, $b_{k,s}$ and $b_{k,e}$ denote the start and end border of $k$-th chunk, respectively.
$\textbf{W}_1^{q}, \textbf{W}_1^{k} \in \mathbb{R}^{d\times d}$ are projection matrices.
If $w_i$ and $w_j$ are not in the same chunk, the attention score between them is set to an infinitely negative value so that the resulting attention score after softmax becomes zero. 

Then we apply softmax on $u_i$ to obtain the normalized scores $\alpha_i$, by which we gain the correlation between $w_i$ and other tokens. 
We aggregate the within-chunk semantics into each token,
\begin{align}
    \tilde{\textbf{h}}_{i}^{w} &= \sum_{j=1}^{M}\alpha_{ij}({\textbf{h}_j^w}\textbf{W}_1^{v}) 
\end{align}
where $\tilde{\textbf{h}}_{i}^{w}$ denotes the updated representation of $\textbf{h}_{i}^w$, $M$ is the number of tokens, $\textbf{W}_1^{v} \in \mathbb{R}^{d\times d}$ is the projection matrix.

The within-chunk semantic interactor explicitly integrates sentence constituents into $\textbf{H}^w$, which helps learn the local semantics.
To model the relationship between image regions, previous works~\cite{scene_graph_textual,graph_image_text_mathing,VLtree} use scene graph parsers to encode regions into visual graphs. We adopt some advanced scene graph parsers approaches~\cite{SGP0,SGP1} but find the dependencies between regions are ambiguous. These fuzzy relationships can cause error propagation, leading to sub-optimal performance.
So we consider each image region as a separate vector, and each region can attend to other regions without limitation. In this way, we integrate the image information into each region representation.

\subsubsection{Cross-chunk Semantic Interactor}

After obtaining the chunk-level semantics via within-chunk semantic interactor, we utilize a cross-chunk semantic interactor to incorporate the global information into each token.
We consider each token as the smallest unit and calculate the attention scores of $w_i$ to the concatenated text-image sequence as follows,
\begin{align}
    u_{ij} &= \frac{({\textbf{h}_{i}^w}{\textbf{W}_2^{q}})(\textbf{h}_j{\textbf{W}_2^{k}})^\mathsf{T}}{\sqrt{d}}
\end{align}
where $\textbf{h}_j$ denotes the representation of the concatenated sequence, $\textbf{W}_2^{q}, \textbf{W}_2^{k} \in \mathbb{R}^{d\times d}$ are projection matrices.

Then we apply softmax on $u_i$ to obtain the normalized scores $\alpha_i$, and aggregate the cross-chunk semantics into each token,
\begin{align}
    \tilde{\textbf{h}}_{i}^{w} =\sum_j\alpha_{ij}{({\textbf{h}_j{\textbf{W}_2^{v}}})} 
\end{align}
where $\textbf{W}_2^{v} \in \mathbb{R}^{d\times d}$ is the projection matrix.

For image representations $\textbf{H}^r$, we consider each region as the smallest unit and update $\textbf{H}^r$ in a similar way.
The cross-chunk semantic interactor helps learn the inter-chunk semantics, fusing the global information in a coarse level.
 
\subsubsection{Cross-modal Semantic Interactor}
Cross-modal semantic interactor aims to conduct semantic vision-language fusion in a fine level.
Unlike previous works~\cite{VQAX,chen2020uniter,oscar} that consider each token separately and build vision-language alignments on token-level, we consider each text chunk as a component to build chunk-level alignments. 
More specifically, we use the average of the tokens in $k$-th chunk as the chunk representation $\textbf{v}_k$,
\begin{align}
    \textbf{v}_k&= \frac{1}{b_{k,e}-b_{k,s}}\sum_{j=b_k^s}^{b_k^e} \textbf{h}_j^w
\label{eqa:v_k}
\end{align}

In this way, we aggregate the semantics within the same chunk. Thereafter, we calculate the relative attention of the $k$-th chunk to the $i$-th region,
\begin{align}
     u_{kj} &= \frac{({\textbf{v}_k}{\textbf{W}_3^{q}})({\textbf{h}_j^r}{\textbf{W}_3^{k}})^\mathsf{T}}{\sqrt{d}}
\end{align}
where $\textbf{h}_j^r$ denotes the representation of $j$-th region, $\textbf{W}_3^{q}, \textbf{W}_3^{k} \in \mathbb{R}^{d\times d}$ are projection matrices.
$u_{kj}$ measures the correlation between $k$-th chunk and $i$-th region, by which we build chunk-level semantic alignments.

We apply softmax on $u_k$ to obtain the normalized scores $\alpha_k$, by which we capture the most salient regions related to the $k$-th chunk.
We aggregate these regions to each token via $\alpha_k$,
\begin{align}
    \tilde{\textbf{h}}_{i}^{w}=\sum_{j=1}^{N}\alpha_{kj}{({\textbf{h}_j^r}{\textbf{W}_3^{v}})}  \quad \text{where $i \in  [b_{k,s},b_{k,e}]$}
\end{align}
where $\textbf{h}_j^r$ denotes the representation of the $j$-th region, $N$ is the number of regions, $\textbf{W}_3^{v} \in \mathbb{R}^{d\times d}$ is projection matrix.

Cross-modal semantic interactor incorporates each chunk and its semantically related regions, which helps remove the ambiguous vision-language alignments and obtain the high-order vision-semantic representations.

\subsection{Relation Inferrer}
\label{Reasoning_module}

We utilize CSI and a pre-trained vision-language model (i.e., Oscar) to obtain comprehensive joint representations of the input pair.
To better retain the alignments of different granularity, we concatenate the outputs of cross-chunk semantic interactor and cross-modal semantic interactor as the final outputs of CSI, represented as $\textbf{O}^C=(\textbf{o}_{CLS}^C,\textbf{o}^{C,w}_1, ..., \textbf{o}^{C,w}_M, \textbf{o}_{SEP}^C,\textbf{o}^{C,r}_g, \textbf{o}^{C,r}_1,...,\textbf{o}^{C,r}_N)$, where $\textbf{o}^{C,w}_i$ and $\textbf{o}^{C,r}_i$ denote the text representations and region representations, respectively.
Similarly, we denote the outputs of Oscar as  $\textbf{O}^T=(\textbf{o}_{CLS}^T,\textbf{o}^{T,w}_1, ..., \textbf{o}^{T,w}_M, \textbf{o}_{SEP}^T,\textbf{o}^{T,r}_g, \textbf{o}^{T,r}_1,...,\textbf{o}^{T,r}_N)$.

$\textbf{o}_{CLS}^T$ and $\textbf{o}_{CLS}^C$ contain coarse-grained holistic vision-language representation on token-level and chunk-level, respectively, which ignores the fine-grained alignments of each token.
To better incorporate the fine-grained vision-language alignments of different levels, we utilize attention mechanism to look back on $\textbf{o}^{T,w}_i$ and $\textbf{o}^{C,w}_i$.
First, we stack $o_{CLS}^T$ and $o_{CLS}^C$, and use a linear projection to keep the dimension unchanged,
\begin{align}
    \textbf{o}_{CLS} = [\textbf{o}_{CLS}^{T};\textbf{o}_{CLS}^{C}]\textbf{W}^{p}
\end{align}
where $\textbf{W}^{p}\in \mathbb{R}^{2d\times d}$ is a projection matrix.

Then we concatenate $\textbf{o}^{T,w}_i$ and $\textbf{o}^{C,w}_i$ to obtain a comprehensive representation $\textbf{O}^w$, which contains token-level and chunk-level cross-modal alignments,
\begin{align}
    \textbf{O}^w &=(\textbf{o}^{T,w}_1, ..., \textbf{o}^{T,w}_M,\textbf{o}^{C,w}_1, ..., \textbf{o}^{C,w}_M)
    \label{o_w}
\end{align}

We calculate the relative score between $\textbf{o}_{CLS}$ and $\textbf{O}^w$,
\begin{align}
    \alpha^I &=\text{softmax}((\textbf{o}_{CLS}{\textbf{W}_4^{q}})(\textbf{O}^w{\textbf{W}_4^{k}})^\mathsf{T}) 
    \label{attn_w}
\end{align}
where $\textbf{W}_4^{q},\textbf{W}_4^{k}  \in \mathbb{R}^{d\times d}$ are projection matrices, $\alpha^I$ denotes the importance of each alignment.
Then we aggregate the salient alignments via $\alpha^I$ and refine $o_{CLS}$,
\begin{align}
    \tilde{\textbf{o}}_{CLS} &=\alpha^I{({\textbf{O}^w{\textbf{W}_4^{v}}})}+\textbf{o}_{CLS}
    \label{refine}
\end{align}
where $\tilde{\textbf{o}}_{CLS}$ denotes the updated $\textbf{o}_{CLS}$, $\textbf{W}_4^{v} \in \mathbb{R}^{d\times d}$ is projection matrix.

We refine $\textbf{o}_{CLS}$ iteratively via Eq.~\ref{attn_w} and Eq.~\ref{refine}.
Finally, a linear classifier is applied to obtain the probability of each relation,
\begin{align}
    \tilde{y}&=\textbf{o}_{CLS}\textbf{W}^{y}
\end{align}
where $\textbf{W}^{y}\in \mathbb{R}^{d\times n}$ is projection matrix, $n$ is the number of relations.
We select the relation with the highest probability as the result.

\subsection{Lexical Constraint-aware Generator}
\label{generator}

Lexical constraint-aware generator (LeCG) aims to generate an explanation $E = (e_1, ..., e_i, ..., e_{L})$ to interpret the decision-making process.
Previous works~\cite{VQAX,evil,GPT_NLE} fail to associate explanation generation with the inference process. To alleviate this problem, LeCG explicitly guides explanation generation with the lexical constraint obtained from the inference process.

First, we adopt a transformer-based~\cite{attn_is_all} language model as the generator.
The generator operates cross-attention over the comprehensive representation $\textbf{O}^w$ (obtained via Eq.~\ref{o_w}) to exploit the input information.
The hidden state $\textbf{h}_t^d$ of the top layer of the generator at time step $t$ is fed to a projection linear and softmax to get the initial generation probability $P_{vocab}(e_t)$ of target token $e_t$.

Then we construct a word set $\mathcal{S}$ as the lexical constraint.
We add up the attention weights $\alpha^I$ (obtained via Eq.~\ref{attn_w}) of each attention layer to get the inference attention score $\alpha^S_i$ of each token, which indicates the importance of each token during inference.
We assume the tokens whose score is higher than the median are essential for the decision-making process, and thus they should be in $\mathcal{S}$,
\begin{equation}
    \mathcal{S}=\{w_i\}  \quad \text{where $\alpha^S_i > \alpha^S_{mid}$}
\end{equation}
where $\alpha^S_i$ is the inference attention score of $i$-th token, and $\alpha^S_{mid}$ is the median score.

We guide the explanation centering around the constraint by combining the initial generation probability $P_{vocab}$ with a lexical constraint probability $P_{lex}$, which is the probability distribution of the tokens within $\mathcal{S}$.
More specifically, we adopt the cross-attention weights of the generator as the score of each input token. We filter those tokens that are not in $\mathcal{S}$ and utilize softmax to get the normalized constrained scores,
\begin{equation}
    {u}^c_{i,t}=\begin{cases}
    \alpha^c_{i,t}& {w}_{i}\in \mathcal{S}\\
    - \infty& {w}_{i}\notin \mathcal{S}
    \end{cases} 
\end{equation}
\begin{equation}
    \tilde{\alpha}^c_{t}=\text{softmax}({u}^c_{t})
\end{equation}
where $\alpha^c_{i,t}$ denotes the original cross-attention score of $w_i$ at time step $t$ from the generator, and $\tilde{\alpha}^c_{t}$ denotes the constrained attention scores of each token.

We sum the $\tilde{\alpha}^c_{i,t}$ where $w_i=e_t$ as the lexical constraint probability of $e_t$,
\begin{equation}
    P_{lex}(e_{t})=\sum_{i:w_i=e_t}\tilde{\alpha}^c_{i,t}
\end{equation}

We use a constrained weight $p_{con}\in [0,1]$ to control the portion of $P_{vocab}(e_t)$ and $P_{lex}(e_t)$ when calculating the final probability.
Following~\cite{pointer,disjoint_vocab}, we calculate the constrained context vector $\textbf{c}_t$,
\begin{align}
    \textbf{c}_t &=\sum_i {\tilde{\alpha}^c_{i,t}}{\textbf{O}^w_i} 
\end{align}

Then we concatenate $\textbf{c}_t$ with the generator output $\textbf{h}^d_t$ and the inputs of language model $\textbf{x}_t$ to obtain $p_{con}$ as follows,
\begin{align}
    p_{con}&=\sigma([\textbf{c}_t;\textbf{h}^d_t;\textbf{x}_{t}]\textbf{W}^{g}) 
\end{align}
where $\sigma(\cdot)$ is a sigmoid activation function, $\textbf{W}^{g}\in\mathbb{R}^{3d\times1}$ is the learning weight matrix.

Last we obtain the final probability of $e_t$ under lexical constraint:
\begin{equation}
    P(e_t)=p_{con}P_{vocab}(e_t)+(1-p_{con})P_{lex}(e_t)
\end{equation}

\subsection{Training and Testing}
\label{training}
\subsubsection{Chunk-aware Semantic Interactor Pre-training}
To improve the accuracy of semantic vision-language alignments, we pre-train CSI on the Flickr30k Entities dataset~\cite{flickrEntitie}.
Flickr30k Entities dataset provides the alignments between noun phrases and image regions, where a phrase is aligned to only one region.
Note that Flickr30k is also the source corpus of e-SNLI-VE~\cite{evil}, so we split the Flickr30k Entities dataset along e-SNLI-VE to avoid data leakage.
During pre-training, we assume that each token should attend to the most semantically relevant region.
We sum the attention weights of the cross-modal semantic interactor layers of $w_i$ and apply softmax on it to get the normalized align score $s_i$. We utilize cross-entropy to enforce the alignment:
\begin{equation}
    \mathcal{L}_{A}=\frac{1}{M} \sum_{i=1}^{M} {(\sum_{j=1}^{N}-z_{ij}\log (\frac{\exp(s_{ij})}{\sum_{k=0}^{N}\exp(s_{ik})}))}
\end{equation}
where $s_{ij}$ is the align score of $i$-th token to $j$-th region, $z_{ij}\in \{0,1\}$ is the label that indicates whether $i$-th token and $j$-th image origin should be aligned (i.e. 1) or not (i.e. 0), $M$ is the number of input tokens and $N$ is the number of image regions.

\subsubsection{Training Pipeline}
The optimization procedure of CALeC contains two stages.
First, we train CSI and the relation inferrer for relation inference until the cross-entropy loss converges:
\begin{equation}
    \mathcal{L}_{Y}=\sum_{i}-y^g_{i}\log (\frac{\exp(\tilde{y}_i)}{\sum_{j}\exp( \tilde{y}_j)})
\end{equation}
where $y^g_{i} \in \{0,1\}$ is the label of the $i$-th relation (e-SNLI-VE) or answer (VQA and VCR), $\tilde{y}_i$ is the 
probability of the $i$-th relation.

Then we freeze the their parameters and train LeCG for explanation generation.
We minimize the negative log-likelihood of LeCG:
\begin{equation}
    \mathcal{L}_{E}=\sum_{i=1}^{L}-\log P(e_i \mid T,I,e_{<i})
\label{equ:loss_g}
\end{equation}
where $L$ denotes the length of the explanation, $e_i$ denotes the target token at time step $t$.

\begin{algorithm}[t]
\caption{Constrained Beam Sample}
\label{algo:constraint-beam}
\begin{algorithmic}[1]
\Require Max explanation length $N$, beam size $k$, sample size $s$,  lexical constraint set $\mathcal{S}$, constraint coefficient $\lambda$.
\Ensure Constrained explanation.
\State $\text{beams} = \text{Generator-Init}(K$)
\For{$1\leq t\leq N$}
    \State probs $=$  Generator-Step(beams);
    \State new\_beams $=$ BeamSample(probs, beams, $k$, $s$)
    \For{$1 \leq i \leq k\times s$}
        \If{$\text{new\_beams}[i].\text{sent}[-1] \in \mathcal{S}$}
            \State $\text{new\_beams}[i].\text{score} = \lambda \times \text{new\_beams}[i].\text{score}$
        \EndIf
    \EndFor
    \State \textbf{Rank} new\_beams \textbf{with} score   
    \State beams $=$ new\_beams[:$k$]
\EndFor
\State \Return beams[0].sent
\end{algorithmic}
\end{algorithm}

\subsubsection{Testing}
To enhance the constraints, we utilize a constrained beam sample during testing.
Conventional beam sample \cite{beam_sample} generates a sentence with the highest probability, which ignores the faithfulness of generated explanation.
To alleviate this problem, we propose a constrained beam sample that scores each beam with the probability and the number of occurrences of the constraint words.
In every step, we multiplied a constraint coefficient $\lambda$ to the candidate who generates a word that is in the lexical constraint set $\mathcal{S}$.
By this way, the candidate who meets more constraints will have a higher score. We choose the candidate with the highest score as the output.
The pseudo-code is in Algorithm~\ref{algo:constraint-beam}.

\section{Experiments}

\subsection{\textbf{Settings}}
\subsubsection{\textbf{Datasets}}

Following the benchmark e-ViL~\cite{evil} for vision-language tasks with NLE, we evaluate our method on the VE-NLE dataset e-SNLI-VE~\cite{evil} and two VQA-NLE datasets VQA-X~\cite{VQAX} and VCR~\cite{VCR}.
\textbf{e-SNLI-VE} is the current biggest VE-NLE dataset that combines SNLI-VE~\cite{vte} and e-SNLI~\cite{e-snli}. The training, validation, and test sets contain 401.7k/14.3k/14.7k image-text pairs, respectively. There are three relations of the input pair: \textit{entailment}, \textit{contradiction} and \textit{neutral}.
\textbf{VQA-X} is a subset of the VQA v2 dataset~\cite{VQA2}, in which each sample contains an image, a question, an answer, and the corresponding explanation. The training, validation, and test sets contain 29.5k/1.5k/2k image-text pairs, respectively.
\textbf{VCR} provides an image, a question and a list of annotated objects. For each question, a model needs to select one answer from four candidates. After that, it needs to select one explanation from four candidates. The test set for VCR is not publicly available. e-ViL~\cite{evil} reorganizes the dataset, and reformulates the explanation selection task as a generation task. The training, validation, and test sets contain 191.6k/21.3k/26.5k image-text pairs, respectively.

\subsubsection{\textbf{Evaluation Metrics}}

Following the e-ViL benchmark, we define three evaluation scores $S_T$, $S_E$, and $S_O$. $S_T$ represents the inference accuracy.
$S_E$ represents the average explanation score of examples inferred correctly. This assumes that an explanation is wrong if it justifies an incorrect answer~\cite{evil}.
We adopt BLEU-4~\cite{BLEU}, ROUGE-L~\cite{ROUGE}, METEOR~\cite{METEOR}, CIDEr~\cite{CIDER} and SPICE~\cite{SPICE} as the explanation scores. All scores are computed with the publicly available code\footnote{https://github.com/tylin/coco-caption}.
$S_O$ represents the overall performance, which is defined as $S_O=S_T \times S_E$.

\subsubsection{\textbf{Baselines}}

Similar to the e-ViL benchmark, we compare our method with five strong baselines.
\textit{Pointing and Justification (PJ-X)}~\cite{VQAX} uses a simplified MCB model \cite{MCB} as the vision-language encoder and an LSTM-based language model as the decoder.
\textit{Faithful Multimodal Explanations (FME)}~\cite{Faithful_NLE} requires the answer and explanation to focus on the same image regions. It utilizes an improved Up-Down VQA
model~\cite{vqa_up_bottom} for answer inference, and an LSTM-based language model for explanation generation.
\textit{Rationale-VT Transformer (RVT)}~\cite{Vision_info_explanation} utilizes different vision-language models to extract vision information and feeds the encoded representations with the question and ground-truth answer to the pre-trained GPT-2~\cite{GPT2}. Note that RVT omits the question answering part, so we directly quoted the results from the e-ViL benchmark, which extends RVT with Bert~\cite{BERT} to obtain the answer. \textit{e-UG}~\cite{evil} combines the powerful pre-trained vision-language model UNITER~\cite{chen2020uniter} and GPT-2.
\textit{NLX-GPT}~\cite{GPT_NLE} utilizes a large-scale pre-trained language model to generate the answer and explanation simultaneously. 

\subsubsection{\textbf{Implementation Details}}

We adopt $\text{Oscar}_{base}$ as the vision-language pre-trained model.
We also utilize its parameters to initialize CSI. The number of layers of within-chunk, cross-chunk, and cross-modal semantic interactors are 3/6/3.
We use a tagging model~\cite{chunkModel} pre-trained on Chunk-CoNLL2000~\cite{cnll2000} to get the text chunk borders. 
The number of attention layers of the relation inferrer is 3.
We adopt $\text{GPT-2}_{base}$~\cite{GPT2} as the transformer-based language model in LeCG and randomly initialize the parameters of cross-attention sub-layers. We regard the input text and answer as prefix information and concatenate them before the explanation.
For training, we use the Adam optimizer~\cite{Adam} with the $10^{-5}$ initial learning rate and linear decay of the learning rate during CSI pre-training and CALeC training pipeline. To maintain the semantic alignment ability of CSI, the initial learning rate of CSI during the training pipeline is set to $10^{-6}$. 
The beam size and top-k of beam sample\footnote{https://huggingface.co/transformers/internal/generation\_utils} are set to 5 and 32.
The constraint coefficient $\lambda$ is set to 0.86.

\subsection{\textbf{Quantitative Analysis}}
\subsubsection{\textbf{Performance Comparison}}

\begin{table*}

  \caption{Automatic evaluation results on the test sets of three datasets.
    B4, R-L, MET are short for BLEU-4, ROUGE-L and METEOR, respectively.
    We directly quote the results of these baselines from their original papers except the ones marked by $^*$, which are obtained by running their released code (e-UG) or evaluating their released output results (NLX-GPT). NLX-GPT evaluates its results using PTBTokenizer~\cite{ptb_toker}, while others do not. For a fair comparison, we provide our $S_E$ w/ (marked by $^\dag$) and w/o PTBTokenizer.
    The best performance is highlighted in bold.}
\renewcommand\arraystretch{0.95} 
\setlength\tabcolsep{11pt}
    \resizebox{0.95\textwidth}{!}{
   \begin{tabular}{@{\quad}c|ccccccccc@{\quad}}
\hline
Dataset & Model  & $S_O$    & $S_T$ & $S_E$ & B4 & R-L & MET. & CIDEr & SPICE \\ \hline
\multirow[m]{7}{*}{e-SNLI-VE} 
& PJ-X \cite{VQAX} & 20.40 & 69.20 & 29.48 & 7.30  & 28.60  & 14.70 & 72.50 & 24.30   \\
& FME\cite{Faithful_NLE}  & 24.19 & 73.70 & 32.82 & 8.20 & \textbf{29.90} & 15.60 & 83.60 & 26.80  \\
& RVT\cite{Vision_info_explanation}  & 24.47 & 72.00 & 33.98 & 9.60 & 27.30 & 18.80 & 81.70 & 32.50   \\
& \multicolumn{1}{c}{e-UG$^*$ \cite{evil} }                  & \multicolumn{1}{c}{27.77}          & \multicolumn{1}{c}{78.28}          & \multicolumn{1}{c}{{\color[HTML]{2B2B2B} 35.48}}          & \multicolumn{1}{c}{10.13}          & \multicolumn{1}{c}{28.09}          & \multicolumn{1}{c}{19.72}          & \multicolumn{1}{c}{85.39}           & 34.07                      \\
& \multicolumn{1}{c}{CALeC}                                  & \multicolumn{1}{c}{\textbf{30.28}} & \multicolumn{1}{c}{\textbf{80.92}} & \multicolumn{1}{c}{{\color[HTML]{2B2B2B} \textbf{37.42}}} & \multicolumn{1}{c}{\textbf{10.53}} & \multicolumn{1}{c}{28.53}          & \multicolumn{1}{c}{\textbf{20.02}} & \multicolumn{1}{c}{\textbf{91.61}}  & \textbf{36.42}             \\ \cdashline{2-10}[1pt/1pt]
& \multicolumn{1}{c}{NLX-GPT$^\dag$  \cite{GPT_NLE}}         & \multicolumn{1}{c}{31.07}          & \multicolumn{1}{c}{73.91}          & \multicolumn{1}{c}{{\color[HTML]{2B2B2B} 42.04}}          & \multicolumn{1}{c}{11.90}          & \multicolumn{1}{c}{33.40}          & \multicolumn{1}{c}{18.10}          & \multicolumn{1}{c}{114.70}          & 32.10                      \\
& \multicolumn{1}{c}{CALeC$^\dag$}                           & \multicolumn{1}{c}{\textbf{37.53}} & \multicolumn{1}{c}{\textbf{80.92}} & \multicolumn{1}{c}{{\color[HTML]{2B2B2B} \textbf{46.38}}} & \multicolumn{1}{c}{\textbf{13.96}} & \multicolumn{1}{c}{\textbf{35.23}} & \multicolumn{1}{c}{\textbf{19.49}} & \multicolumn{1}{c}{\textbf{127.22}} & \textbf{35.98}             \\ \hline  \hline
\multirow[m]{7}{*}{VQA-X}
& \multicolumn{1}{c}{PJ-X \cite{VQAX}}                      & \multicolumn{1}{c}{28.76}          & \multicolumn{1}{c}{76.40}          & \multicolumn{1}{c}{37.64}                                 & \multicolumn{1}{c}{22.70}          & \multicolumn{1}{c}{46.00}          & \multicolumn{1}{c}{19.70}          & \multicolumn{1}{c}{82.70}           & 17.10                      \\
& \multicolumn{1}{c}{FME \cite{Faithful_NLE} }              & \multicolumn{1}{c}{29.60}          & \multicolumn{1}{c}{75.50}          & \multicolumn{1}{c}{39.20}                                 & \multicolumn{1}{c}{23.10}          & \multicolumn{1}{c}{\textbf{47.10}} & \multicolumn{1}{c}{20.40}          & \multicolumn{1}{c}{\textbf{87.00}}  & 18.40                      \\
& \multicolumn{1}{c}{RVT \cite{Vision_info_explanation} }   & \multicolumn{1}{c}{20.17}          & \multicolumn{1}{c}{68.60}          & \multicolumn{1}{c}{29.40}                                 & \multicolumn{1}{c}{17.40}          & \multicolumn{1}{c}{42.10}          & \multicolumn{1}{c}{19.20}          & \multicolumn{1}{c}{52.50}           & 15.80                      \\
& \multicolumn{1}{c}{e-UG \cite{evil}}                      & \multicolumn{1}{c}{29.82}          & \multicolumn{1}{c}{80.50}          & \multicolumn{1}{c}{37.04}                                 & \multicolumn{1}{c}{23.20}          & \multicolumn{1}{c}{45.70}          & \multicolumn{1}{c}{22.10}          & \multicolumn{1}{c}{74.10}           & 20.10                      \\
& \multicolumn{1}{c}{CALeC}                                 & \multicolumn{1}{c}{\textbf{34.43}} & \multicolumn{1}{c}{\textbf{86.38}} & \multicolumn{1}{c}{\textbf{39.85}}                        & \multicolumn{1}{c}{\textbf{25.47}} & \multicolumn{1}{c}{47.02}          & \multicolumn{1}{c}{\textbf{23.38}} & \multicolumn{1}{c}{81.58}           & \textbf{21.82}             \\ \cdashline{2-10}[1pt/1pt]
& \multicolumn{1}{c}{NLX-GPT$^\dag$ \cite{GPT_NLE}}         & \multicolumn{1}{c}{39.18}          & \multicolumn{1}{c}{83.07}          & \multicolumn{1}{c}{47.16}                                 & \multicolumn{1}{c}{28.50}          & \multicolumn{1}{c}{51.50}          & \multicolumn{1}{c}{\textbf{23.10}} & \multicolumn{1}{c}{110.60}          & \textbf{22.10}             \\
& \multicolumn{1}{c}{CALeC$^\dag$}                          & \multicolumn{1}{c}{\textbf{40.87}} & \multicolumn{1}{c}{\textbf{86.38}} & \multicolumn{1}{c}{\textbf{47.31}}                        & \multicolumn{1}{c}{\textbf{29.30}} & \multicolumn{1}{c}{\textbf{51.59}} & \multicolumn{1}{c}{23.07}          & \multicolumn{1}{c}{\textbf{110.90}} & 21.69                      \\ \hline  \hline
\multirow[m]{7}{*}{VCR}
& \multicolumn{1}{c}{PJ-X \cite{VQAX}}                      & \multicolumn{1}{c}{4.98}           & \multicolumn{1}{c}{39.00}          & \multicolumn{1}{c}{{\color[HTML]{2B2B2B} 12.76}}          & \multicolumn{1}{c}{3.40}           & \multicolumn{1}{c}{20.50}          & \multicolumn{1}{c}{16.40}          & \multicolumn{1}{c}{19.00}           & 4.50                       \\
& \multicolumn{1}{c}{FME \cite{Faithful_NLE}}               & \multicolumn{1}{c}{9.42}           & \multicolumn{1}{c}{48.90}          & \multicolumn{1}{c}{{\color[HTML]{2B2B2B} \textbf{19.26}}}          & \multicolumn{1}{c}{4.40}           & \multicolumn{1}{c}{22.70}          & \multicolumn{1}{c}{\textbf{17.30}}          & \multicolumn{1}{c}{27.70}           & \textbf{24.20}                      \\
& \multicolumn{1}{c}{RVT \cite{Vision_info_explanation}}    & \multicolumn{1}{c}{9.29}           & \multicolumn{1}{c}{59.00}          & \multicolumn{1}{c}{{\color[HTML]{2B2B2B} 15.74}}          & \multicolumn{1}{c}{3.80}           & \multicolumn{1}{c}{21.90}          & \multicolumn{1}{c}{11.20}          & \multicolumn{1}{c}{30.10}           & 11.70                      \\
& \multicolumn{1}{c}{e-UG \cite{evil}}                      & \multicolumn{1}{c}{11.71}          & \multicolumn{1}{c}{69.80}          & \multicolumn{1}{c}{{\color[HTML]{2B2B2B} 16.78}}          & \multicolumn{1}{c}{4.30}           & \multicolumn{1}{c}{22.50}          & \multicolumn{1}{c}{11.80}          & \multicolumn{1}{c}{32.70}           & 12.60                      \\
& \multicolumn{1}{c}{CALeC}                                 & \multicolumn{1}{c}{\textbf{13.95}}               & \multicolumn{1}{c}{\textbf{73.03}} & \multicolumn{1}{c}{19.10}                                      & \multicolumn{1}{c}{\textbf{5.59}}               & \multicolumn{1}{c}{\textbf{22.99}}               & \multicolumn{1}{c}{12.78}               & \multicolumn{1}{c}{\textbf{39.61}}                &   \multicolumn{1}{c}{14.54}                   \\ \cdashline{2-10}[1pt/1pt]
& \multicolumn{1}{c}{NLX-GPT$^{*\dag}$ \cite{GPT_NLE}}         & \multicolumn{1}{c}{1.88}             & \multicolumn{1}{c}{13.45}             & \multicolumn{1}{c}{{\color[HTML]{2B2B2B} 13.96}}          & \multicolumn{1}{c}{3.16}           & \multicolumn{1}{c}{20.76}          & \multicolumn{1}{c}{8.62}          & \multicolumn{1}{c}{27.72}           & \multicolumn{1}{c}{9.54} \\
& \multicolumn{1}{c}{CALeC$^\dag$}                          & \multicolumn{1}{c}{\textbf{15.70}}             & \multicolumn{1}{c}{\textbf{73.03}} & \multicolumn{1}{c}{\textbf{21.50}}                                      & \multicolumn{1}{c}{\textbf{6.34}}               & \multicolumn{1}{c}{\textbf{25.22}}               & \multicolumn{1}{c}{\textbf{12.22}}               & \multicolumn{1}{c}{\textbf{49.35}}                &  \multicolumn{1}{c}{\textbf{14.37}} \\ 
\hline
\end{tabular}
    }
    \label{tab:exp_result}
\end{table*}

We compare our proposed method CALeC against five strong methods on three datasets. The automatic evaluation results are shown in Table~\ref{tab:exp_result}. 
We can see that CALeC achieves the best performance, substantially surpassing all the baseline on $S_O$.
By effectively performing chunk-aware semantic alignment and conducting inference over the fine-grained vision-language alignments, CALeC outperforms the strongest baseline model by $2.64$, $3.31$, and $3.23$ points on $S_T$ metric across the three datasets, respectively.
Though the three datasets focus on different vision-language tasks, CALeC gains accuracy improvement all over them. It suggests that building accurate semantic alignment is a common yet crucial backbone for vision-language models.
CALeC also surpasses the state-of-the-art model NLX-GPT on $S_E$ of the three datasets. This verifies that explicitly guiding the generator through lexical constraint can help improve the quality of generated explanations.
We observe that $S_E$ of FME in VCR is slightly higher than CALeC. This may be attributed to the lower $S_T$ of FME, so FME only needs to count the more accessible samples when calculating $S_E$.
Note that NLX-GPT does not provide its inference accuracy on the VCR dataset, so we calculate the scores based on their released output results\footnote{https://github.com/fawazsammani/nlxgpt}.
The $S_T$ of NLX-GPT is exceptionally low in VCR. This probably because that the answer of VCR is much longer than the other datasets, so it is harder for NLX-GPT to generate the correct answer.

\subsubsection{\textbf{Ablation Study}}

\begin{table}
\caption{Ablation studies of CALeC on the test sets. CBS, RI are short for constrained beam sample and relation inferrer.
}
\centering
    \resizebox{\linewidth}{!}{
        \begin{tabular}{ccccc}
        \hline
                Model                        & Overall & e-SNLI-VE       & VQA-X          & VCR  \\ \hline
                CALeC                   &  \textbf{31.37}       & \textbf{37.53}  &\textbf{40.87}  & \textbf{15.70}    \\
                w/o CBS         &  30.72$^{\downarrow 0.65}$       & 37.18           & 39.98         &  14.99    \\
                w/o LeCG          &  30.23$^{\downarrow 1.14}$       & 36.70           & 38.76          & 15.23     \\
                w/o LeCG \& CBS  &  29.68$^{\downarrow 1.69}$       & 35.78           & 38.58          & 14.67 \\
                w/o RI \& LeCG \& CBS  &  28.78$^{\downarrow 2.59}$       & 35.29           & 36.58          & 14.48 \\
                w/o CSI \& RI \& LeCG \& CBS      & 27.48$^{\downarrow 3.89}$   & 34.50           & 33.81         & 14.12 \\ \hline
        \end{tabular}
    }
    \label{tab:ablation_study}
\end{table}

We conduct ablation experiments to verify the effectiveness of CSI, relation inferrer, and LeCG in CALeC, which are presented in Table~\ref{tab:ablation_study}. We only list $S_O$ because it  summarizes the performance on both $S_T$ and $S_E$. For a fair comparison, all the evaluated models have the same experimental settings and generate explanations through the beam sample algorithm. 
The second line verifies the effectiveness of the constrained beam sample. We can see that adding constraints to the conventional beam sample algorithm can help improve the quality of generated explanations.
The third line shows the results when we drop LeCG and only retain the transformer-based generator. LeCG has a more significant influence than the constrained beam sample, indicating that directly guiding the generator with constraints can perform better than the post-hoc edit method.
When we drop LeCG and the constrained beam sample simultaneously, the decrease in the overall score (1.69) is almost equal to the sum of separate reductions (1.79). This phenomenon shows that these two constrained approaches act at different but complementary points during generation and can jointly improve the quality of explanations.
The constraint set is formed based on the relation inferrer and CSI, so they cannot be dropped solely. We drop the relation inferrer along with the constrained methods, in which we directly utilize the linear classifier on the concatenation of the two [CLS] outputs. The scores on the three datasets all decrease, indicating that the relation inferrer can better incorporate the fine-grained alignments of different level.
We then drop CSI along with other components, in which the model degenerates into the vanilla transformer-based seq2seq model, i.e., Oscar-GPT.
There is a 1.39 net decrease compared to just dropping the relation inferrer, which is higher than other components. This result shows that chunk-aware semantic alignment can greatly benefit vision-language tasks with NLE.

\subsection{\textbf{Qualitative Analysis}}

\subsubsection{\textbf{Human Evaluation}}
\begin{figure}[]
  \centering
  \includegraphics[width=\linewidth]{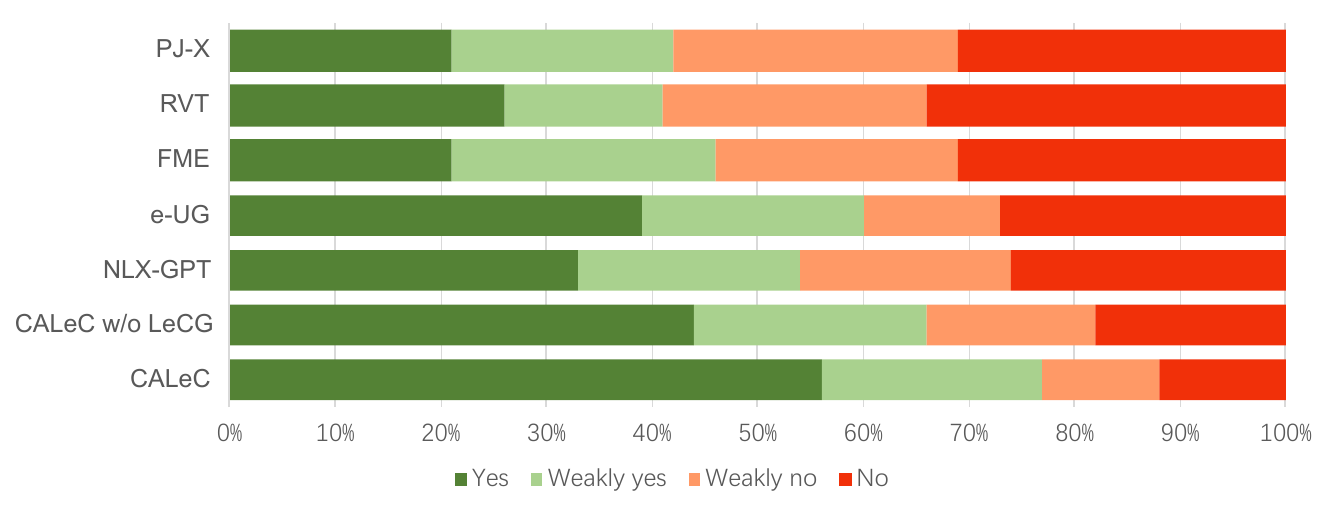}
  \caption{Human evaluation results on e-SNLI-VE.}
  \label{fig:human_eval}
\end{figure}

The automatic NLG metrics do not always reflect the faithfulness of the explanations because explanations can come in different forms and be very generic and data-biased. So we adopt human evaluation to evaluate the faithfulness of explanations.
We conduct human evaluation on the test set of e-SNLI-VE, because we do not find the public result of the baseline models on other datasets. Following the e-ViL benchmark, we randomly select 100 test samples with correctly predicted answers. We ask annotators ``Given the image and the hypothesis, does the explanation justify the answer?'' with four choices: \textit{Yes}, \textit{Weakly yes}, \textit{Weakly no} and \textit{No}.
To ensure the fairness of assessment, the explanations of each sample are shuffled.
As shown in the last bar of Figure~\ref{fig:human_eval}, CALeC gets about 77\% positive scores (green region), and about 55\% of them are strongly positive (dark green region), which far surpasses other models. The results indicate that our explanations can justify the answer better and reflect the inference process faithfully. We also conduct a human evaluation on CALeC w/o LeCG (the next-to-last bar). The proportion of \textit{Yes} obviously decreases and the proportion of negative choices increases. This phenomenon verifies that adding constraints on explanation generation can guide the generator to focus on the input and generate explanations faithful to the inference process.

\subsubsection{Case Study.}
\begin{figure}[]
  \centering
  \includegraphics[width=0.9\linewidth]{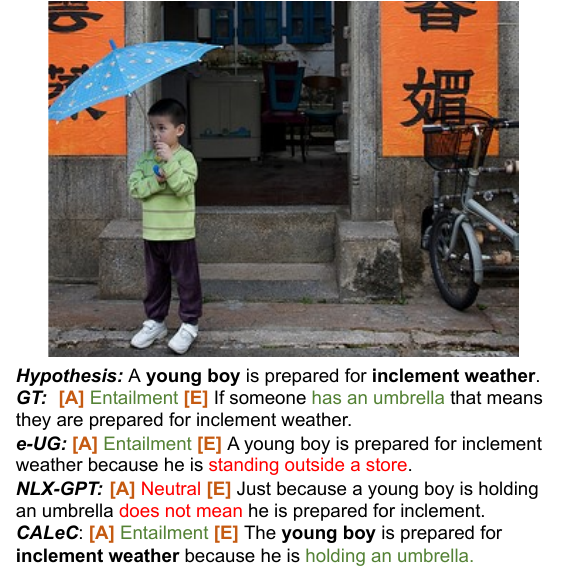}
  \caption{An example on e-SNLI-VE. Bold words are chose as the lexical constraint.}
  \label{fig:compare_case}
\end{figure}

\begin{figure}[]
  \centering
  \includegraphics[width=\linewidth]{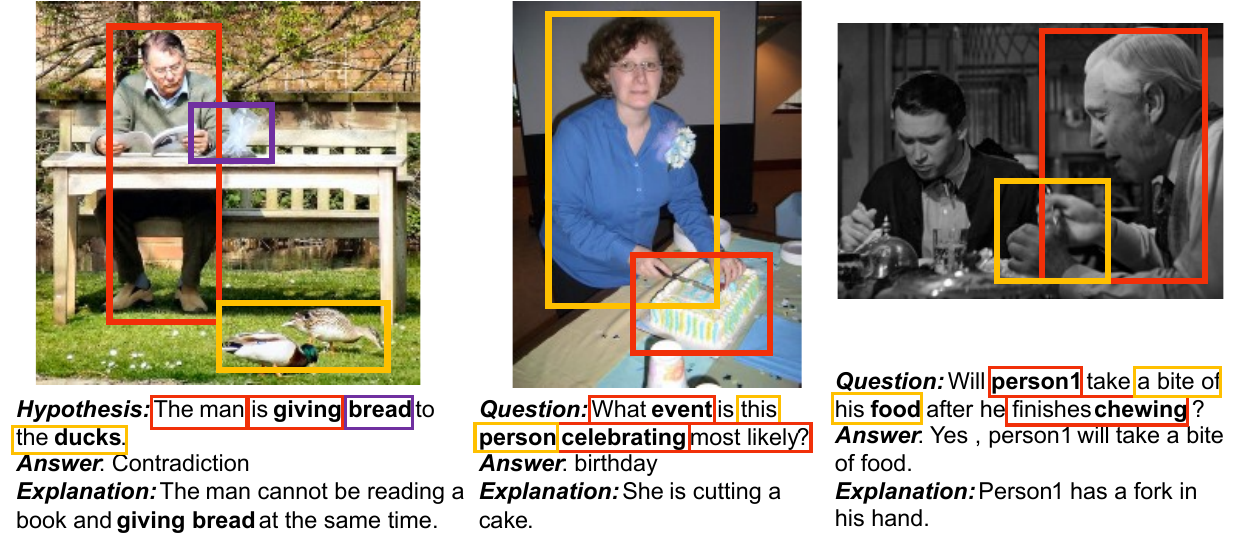}
  \caption{Examples from our model on e-SNLI-VE, VQA-X and VCR. The same colors show the alignments between chunks and image regions. Bold words are the lexical constraint.
}
  \label{fig:cases}
\end{figure}

In Figure~\ref{fig:compare_case}, we show an example with the inference result and explanations of each model on e-SNLI-VE.
In this example, CALeC is the only model that infers the correct relation and generates a faithful explanation. In contrast, e-UG mistakes \textit{a house} for \textit{a shop} and generates an illogical explanation, and NLX-GPT predicts the wrong answer.
In Figure~\ref{fig:cases}, we show some qualitative results from our model on the three datasets. 
Based on the semantic alignments, the relation inferrer can accurately find the keywords (bold words in input text).
LeCG can generate faithful explanations relevant to the inference process and input pair.
We observe that although we only provide alignments for noun chunks during pre-training for CSI, it can learn alignments for other part-of-speech chunks (e.g. \textit{is giving}) during fine-tuning, which may benefit from the cross-chunk semantic interactor.

\section{Conclusion and Future Directions}

We present a unified Chunk-aware Alignment and Lexical Constraint based method (CALeC) for Visual Entailment with Natural Language Explanations (VE-NLE). Our work is motivated by the need to exploit the rich semantics contained in the chunks and generate explanations faithful to the inference process.
This method builds chunk-aware semantic alignment and incorporates the keywords of the inference process into explanation to enhance faithfulness.
We conduct extensive experiments on three datasets. Experimental results show that our method achieves state-of-the-art performance on relation inference and explanation generation. It also has strong generalizability over other vision-language tasks.
Future work includes building alignments between chunks and visual concepts rather than predetermined regions and improving the relevance between explanations and input image.

\begin{acks}
This work is jointly supported by grants: Natural Science Foundation of China (No. 62006061 and 61872107), Stable Support Program for Higher Education Institutions of Shenzhen (No. GXWD20201230 155427003-20200824155011001) and Strategic Emerging Industry Development Special Funds of Shenzhen(No. JCYJ20200109113441941).

\end{acks}
\clearpage
\bibliographystyle{ACM-Reference-Format}
\bibliography{acmm}


\begin{thebibliography}{46}


\ifx \showCODEN    \undefined \def \showCODEN     #1{\unskip}     \fi
\ifx \showDOI      \undefined \def \showDOI       #1{#1}\fi
\ifx \showISBNx    \undefined \def \showISBNx     #1{\unskip}     \fi
\ifx \showISBNxiii \undefined \def \showISBNxiii  #1{\unskip}     \fi
\ifx \showISSN     \undefined \def \showISSN      #1{\unskip}     \fi
\ifx \showLCCN     \undefined \def \showLCCN      #1{\unskip}     \fi
\ifx \shownote     \undefined \def \shownote      #1{#1}          \fi
\ifx \showarticletitle \undefined \def \showarticletitle #1{#1}   \fi
\ifx \showURL      \undefined \def \showURL       {\relax}        \fi
\providecommand\bibfield[2]{#2}
\providecommand\bibinfo[2]{#2}
\providecommand\natexlab[1]{#1}
\providecommand\showeprint[2][]{arXiv:#2}

\bibitem[Anderson et~al\mbox{.}(2016)]%
        {SPICE}
\bibfield{author}{\bibinfo{person}{Peter Anderson}, \bibinfo{person}{Basura
  Fernando}, \bibinfo{person}{Mark Johnson}, {and} \bibinfo{person}{Stephen
  Gould}.} \bibinfo{year}{2016}\natexlab{}.
\newblock \showarticletitle{Spice: Semantic propositional image caption
  evaluation}. In \bibinfo{booktitle}{\emph{European conference on computer
  vision}}. Springer, \bibinfo{pages}{382--398}.
\newblock


\bibitem[Anderson et~al\mbox{.}(2018)]%
        {vqa_up_bottom}
\bibfield{author}{\bibinfo{person}{Peter Anderson}, \bibinfo{person}{Xiaodong
  He}, \bibinfo{person}{Chris Buehler}, \bibinfo{person}{Damien Teney},
  \bibinfo{person}{Mark Johnson}, \bibinfo{person}{Stephen Gould}, {and}
  \bibinfo{person}{Lei Zhang}.} \bibinfo{year}{2018}\natexlab{}.
\newblock \showarticletitle{Bottom-up and top-down attention for image
  captioning and visual question answering}. In
  \bibinfo{booktitle}{\emph{Proceedings of the IEEE conference on computer
  vision and pattern recognition}}. \bibinfo{pages}{6077--6086}.
\newblock


\bibitem[Banerjee and Lavie(2005)]%
        {METEOR}
\bibfield{author}{\bibinfo{person}{Satanjeev Banerjee} {and}
  \bibinfo{person}{Alon Lavie}.} \bibinfo{year}{2005}\natexlab{}.
\newblock \showarticletitle{{METEOR:} An Automatic Metric for {MT} Evaluation
  with Improved Correlation with Human Judgments}. In
  \bibinfo{booktitle}{\emph{Proceedings of the Workshop on Intrinsic and
  Extrinsic Evaluation Measures for Machine Translation and/or
  Summarization@ACL 2005, Ann Arbor, Michigan, USA, June 29, 2005}},
  \bibfield{editor}{\bibinfo{person}{Jade Goldstein}, \bibinfo{person}{Alon
  Lavie}, \bibinfo{person}{Chin{-}Yew Lin}, {and} \bibinfo{person}{Clare~R.
  Voss}} (Eds.). \bibinfo{publisher}{Association for Computational
  Linguistics}, \bibinfo{pages}{65--72}.
\newblock


\bibitem[Bowman et~al\mbox{.}(2015)]%
        {snli}
\bibfield{author}{\bibinfo{person}{Samuel~R. Bowman}, \bibinfo{person}{Gabor
  Angeli}, \bibinfo{person}{Christopher Potts}, {and}
  \bibinfo{person}{Christopher~D. Manning}.} \bibinfo{year}{2015}\natexlab{}.
\newblock \showarticletitle{A large annotated corpus for learning natural
  language inference}. In \bibinfo{booktitle}{\emph{Proceedings of the 2015
  Conference on Empirical Methods in Natural Language Processing}}.
  \bibinfo{publisher}{Association for Computational Linguistics},
  \bibinfo{address}{Lisbon, Portugal}, \bibinfo{pages}{632--642}.
\newblock


\bibitem[Camburu et~al\mbox{.}(2018)]%
        {e-snli}
\bibfield{author}{\bibinfo{person}{Oana-Maria Camburu}, \bibinfo{person}{Tim
  Rockt{\"a}schel}, \bibinfo{person}{Thomas Lukasiewicz}, {and}
  \bibinfo{person}{Phil Blunsom}.} \bibinfo{year}{2018}\natexlab{}.
\newblock \showarticletitle{e-snli: Natural language inference with natural
  language explanations}.
\newblock \bibinfo{journal}{\emph{Advances in Neural Information Processing
  Systems}}  \bibinfo{volume}{31} (\bibinfo{year}{2018}).
\newblock


\bibitem[Chen et~al\mbox{.}(2020)]%
        {chen2020uniter}
\bibfield{author}{\bibinfo{person}{Yen-Chun Chen}, \bibinfo{person}{Linjie Li},
  \bibinfo{person}{Licheng Yu}, \bibinfo{person}{Ahmed El~Kholy},
  \bibinfo{person}{Faisal Ahmed}, \bibinfo{person}{Zhe Gan},
  \bibinfo{person}{Yu Cheng}, {and} \bibinfo{person}{Jingjing Liu}.}
  \bibinfo{year}{2020}\natexlab{}.
\newblock \showarticletitle{Uniter: Universal image-text representation
  learning}. In \bibinfo{booktitle}{\emph{European conference on computer
  vision}}. Springer, \bibinfo{pages}{104--120}.
\newblock


\bibitem[Devlin et~al\mbox{.}(2018)]%
        {BERT}
\bibfield{author}{\bibinfo{person}{Jacob Devlin}, \bibinfo{person}{Ming-Wei
  Chang}, \bibinfo{person}{Kenton Lee}, {and} \bibinfo{person}{Kristina
  Toutanova}.} \bibinfo{year}{2018}\natexlab{}.
\newblock \showarticletitle{Bert: Pre-training of deep bidirectional
  transformers for language understanding}.
\newblock \bibinfo{journal}{\emph{arXiv preprint arXiv:1810.04805}}
  (\bibinfo{year}{2018}).
\newblock


\bibitem[Dua et~al\mbox{.}(2021)]%
        {exlpanation_refine}
\bibfield{author}{\bibinfo{person}{Radhika Dua}, \bibinfo{person}{Sai~Srinivas
  Kancheti}, {and} \bibinfo{person}{Vineeth~N. Balasubramanian}.}
  \bibinfo{year}{2021}\natexlab{}.
\newblock \showarticletitle{Beyond {VQA:} Generating Multi-Word Answers and
  Rationales to Visual Questions}. In \bibinfo{booktitle}{\emph{{IEEE}
  Conference on Computer Vision and Pattern Recognition Workshops, {CVPR}
  Workshops 2021, virtual, June 19-25, 2021}}. \bibinfo{publisher}{Computer
  Vision Foundation / {IEEE}}, \bibinfo{pages}{1623--1632}.
\newblock


\bibitem[Fukui et~al\mbox{.}(2016)]%
        {MCB}
\bibfield{author}{\bibinfo{person}{Akira Fukui}, \bibinfo{person}{Dong~Huk
  Park}, \bibinfo{person}{Daylen Yang}, \bibinfo{person}{Anna Rohrbach},
  \bibinfo{person}{Trevor Darrell}, {and} \bibinfo{person}{Marcus Rohrbach}.}
  \bibinfo{year}{2016}\natexlab{}.
\newblock \showarticletitle{Multimodal Compact Bilinear Pooling for Visual
  Question Answering and Visual Grounding}. In
  \bibinfo{booktitle}{\emph{Proceedings of the 2016 Conference on Empirical
  Methods in Natural Language Processing, {EMNLP} 2016, Austin, Texas, USA,
  November 1-4, 2016}}, \bibfield{editor}{\bibinfo{person}{Jian Su},
  \bibinfo{person}{Xavier Carreras}, {and} \bibinfo{person}{Kevin Duh}} (Eds.).
  \bibinfo{publisher}{The Association for Computational Linguistics},
  \bibinfo{pages}{457--468}.
\newblock
\urldef\tempurl%
\url{https://doi.org/10.18653/v1/d16-1044}
\showDOI{\tempurl}


\bibitem[Ge et~al\mbox{.}(2021)]%
        {VLtree}
\bibfield{author}{\bibinfo{person}{Xuri Ge}, \bibinfo{person}{Fuhai Chen},
  \bibinfo{person}{Joemon~M. Jose}, \bibinfo{person}{Zhilong Ji},
  \bibinfo{person}{Zhongqin Wu}, {and} \bibinfo{person}{Xiao Liu}.}
  \bibinfo{year}{2021}\natexlab{}.
\newblock \showarticletitle{Structured Multi-modal Feature Embedding and
  Alignment for Image-Sentence Retrieval}. In \bibinfo{booktitle}{\emph{{MM}
  '21: {ACM} Multimedia Conference, Virtual Event, China, October 20 - 24,
  2021}}, \bibfield{editor}{\bibinfo{person}{Heng~Tao Shen},
  \bibinfo{person}{Yueting Zhuang}, \bibinfo{person}{John~R. Smith},
  \bibinfo{person}{Yang Yang}, \bibinfo{person}{Pablo Cesar},
  \bibinfo{person}{Florian Metze}, {and} \bibinfo{person}{Balakrishnan
  Prabhakaran}} (Eds.). \bibinfo{publisher}{{ACM}},
  \bibinfo{pages}{5185--5193}.
\newblock


\bibitem[Goyal et~al\mbox{.}(2017)]%
        {VQA2}
\bibfield{author}{\bibinfo{person}{Yash Goyal}, \bibinfo{person}{Tejas Khot},
  \bibinfo{person}{Douglas Summers{-}Stay}, \bibinfo{person}{Dhruv Batra},
  {and} \bibinfo{person}{Devi Parikh}.} \bibinfo{year}{2017}\natexlab{}.
\newblock \showarticletitle{Making the {V} in {VQA} Matter: Elevating the Role
  of Image Understanding in Visual Question Answering}. In
  \bibinfo{booktitle}{\emph{2017 {IEEE} Conference on Computer Vision and
  Pattern Recognition, {CVPR} 2017, Honolulu, HI, USA, July 21-26, 2017}}.
  \bibinfo{publisher}{{IEEE} Computer Society}, \bibinfo{pages}{6325--6334}.
\newblock


\bibitem[Holtzman et~al\mbox{.}(2020)]%
        {beam_sample}
\bibfield{author}{\bibinfo{person}{Ari Holtzman}, \bibinfo{person}{Jan Buys},
  \bibinfo{person}{Li Du}, \bibinfo{person}{Maxwell Forbes}, {and}
  \bibinfo{person}{Yejin Choi}.} \bibinfo{year}{2020}\natexlab{}.
\newblock \showarticletitle{The Curious Case of Neural Text Degeneration}. In
  \bibinfo{booktitle}{\emph{8th International Conference on Learning
  Representations, {ICLR} 2020, Addis Ababa, Ethiopia, April 26-30, 2020}}.
\newblock


\bibitem[Huang et~al\mbox{.}(2021)]%
        {soho}
\bibfield{author}{\bibinfo{person}{Zhicheng Huang}, \bibinfo{person}{Zhaoyang
  Zeng}, \bibinfo{person}{Yupan Huang}, \bibinfo{person}{Bei Liu},
  \bibinfo{person}{Dongmei Fu}, {and} \bibinfo{person}{Jianlong Fu}.}
  \bibinfo{year}{2021}\natexlab{}.
\newblock \showarticletitle{Seeing out of the box: End-to-end pre-training for
  vision-language representation learning}. In
  \bibinfo{booktitle}{\emph{Proceedings of the IEEE/CVF Conference on Computer
  Vision and Pattern Recognition}}. \bibinfo{pages}{12976--12985}.
\newblock


\bibitem[Kayser et~al\mbox{.}(2021)]%
        {evil}
\bibfield{author}{\bibinfo{person}{Maxime Kayser}, \bibinfo{person}{Oana-Maria
  Camburu}, \bibinfo{person}{Leonard Salewski}, \bibinfo{person}{Cornelius
  Emde}, \bibinfo{person}{Virginie Do}, \bibinfo{person}{Zeynep Akata}, {and}
  \bibinfo{person}{Thomas Lukasiewicz}.} \bibinfo{year}{2021}\natexlab{}.
\newblock \showarticletitle{e-vil: A dataset and benchmark for natural language
  explanations in vision-language tasks}. In
  \bibinfo{booktitle}{\emph{Proceedings of the IEEE/CVF International
  Conference on Computer Vision}}. \bibinfo{pages}{1244--1254}.
\newblock


\bibitem[Kingma and Ba(2015)]%
        {Adam}
\bibfield{author}{\bibinfo{person}{Diederik~P. Kingma} {and}
  \bibinfo{person}{Jimmy Ba}.} \bibinfo{year}{2015}\natexlab{}.
\newblock \showarticletitle{Adam: {A} Method for Stochastic Optimization}. In
  \bibinfo{booktitle}{\emph{3rd International Conference on Learning
  Representations, {ICLR} 2015, San Diego, CA, USA, May 7-9, 2015, Conference
  Track Proceedings}}, \bibfield{editor}{\bibinfo{person}{Yoshua Bengio} {and}
  \bibinfo{person}{Yann LeCun}} (Eds.).
\newblock
\urldef\tempurl%
\url{http://arxiv.org/abs/1412.6980}
\showURL{%
\tempurl}


\bibitem[Li et~al\mbox{.}(2020)]%
        {oscar}
\bibfield{author}{\bibinfo{person}{Xiujun Li}, \bibinfo{person}{Xi Yin},
  \bibinfo{person}{Chunyuan Li}, \bibinfo{person}{Pengchuan Zhang},
  \bibinfo{person}{Xiaowei Hu}, \bibinfo{person}{Lei Zhang},
  \bibinfo{person}{Lijuan Wang}, \bibinfo{person}{Houdong Hu},
  \bibinfo{person}{Li Dong}, \bibinfo{person}{Furu Wei}, {et~al\mbox{.}}}
  \bibinfo{year}{2020}\natexlab{}.
\newblock \showarticletitle{Oscar: Object-semantics aligned pre-training for
  vision-language tasks}. In \bibinfo{booktitle}{\emph{European Conference on
  Computer Vision}}. Springer, \bibinfo{pages}{121--137}.
\newblock


\bibitem[Lin(2004)]%
        {ROUGE}
\bibfield{author}{\bibinfo{person}{Chin-Yew Lin}.}
  \bibinfo{year}{2004}\natexlab{}.
\newblock \showarticletitle{Rouge: A package for automatic evaluation of
  summaries}. In \bibinfo{booktitle}{\emph{Text summarization branches out}}.
  \bibinfo{pages}{74--81}.
\newblock


\bibitem[Lin et~al\mbox{.}(2014)]%
        {mscoco}
\bibfield{author}{\bibinfo{person}{Tsung-Yi Lin}, \bibinfo{person}{Michael
  Maire}, \bibinfo{person}{Serge Belongie}, \bibinfo{person}{James Hays},
  \bibinfo{person}{Pietro Perona}, \bibinfo{person}{Deva Ramanan},
  \bibinfo{person}{Piotr Doll{\'a}r}, {and} \bibinfo{person}{C~Lawrence
  Zitnick}.} \bibinfo{year}{2014}\natexlab{}.
\newblock \showarticletitle{Microsoft coco: Common objects in context}. In
  \bibinfo{booktitle}{\emph{European conference on computer vision}}. Springer,
  \bibinfo{pages}{740--755}.
\newblock


\bibitem[Liu et~al\mbox{.}(2020)]%
        {graph_image_text_mathing}
\bibfield{author}{\bibinfo{person}{Chunxiao Liu}, \bibinfo{person}{Zhendong
  Mao}, \bibinfo{person}{Tianzhu Zhang}, \bibinfo{person}{Hongtao Xie},
  \bibinfo{person}{Bin Wang}, {and} \bibinfo{person}{Yongdong Zhang}.}
  \bibinfo{year}{2020}\natexlab{}.
\newblock \showarticletitle{Graph structured network for image-text matching}.
  In \bibinfo{booktitle}{\emph{Proceedings of the IEEE/CVF Conference on
  Computer Vision and Pattern Recognition}}. \bibinfo{pages}{10921--10930}.
\newblock


\bibitem[Manning et~al\mbox{.}(2014)]%
        {ptb_toker}
\bibfield{author}{\bibinfo{person}{Christopher~D. Manning},
  \bibinfo{person}{Mihai Surdeanu}, \bibinfo{person}{John Bauer},
  \bibinfo{person}{Jenny~Rose Finkel}, \bibinfo{person}{Steven Bethard}, {and}
  \bibinfo{person}{David McClosky}.} \bibinfo{year}{2014}\natexlab{}.
\newblock \showarticletitle{The Stanford CoreNLP Natural Language Processing
  Toolkit}. In \bibinfo{booktitle}{\emph{Proceedings of the 52nd Annual Meeting
  of the Association for Computational Linguistics, {ACL} 2014, June 22-27,
  2014, Baltimore, MD, USA, System Demonstrations}}. \bibinfo{publisher}{The
  Association for Computer Linguistics}, \bibinfo{pages}{55--60}.
\newblock
\urldef\tempurl%
\url{https://doi.org/10.3115/v1/p14-5010}
\showDOI{\tempurl}


\bibitem[Marasovi{\'c} et~al\mbox{.}(2020)]%
        {Vision_info_explanation}
\bibfield{author}{\bibinfo{person}{Ana Marasovi{\'c}}, \bibinfo{person}{Chandra
  Bhagavatula}, \bibinfo{person}{Jae~sung Park}, \bibinfo{person}{Ronan
  Le~Bras}, \bibinfo{person}{Noah~A. Smith}, {and} \bibinfo{person}{Yejin
  Choi}.} \bibinfo{year}{2020}\natexlab{}.
\newblock \showarticletitle{Natural Language Rationales with Full-Stack Visual
  Reasoning: From Pixels to Semantic Frames to Commonsense Graphs}. In
  \bibinfo{booktitle}{\emph{Findings of the Association for Computational
  Linguistics: EMNLP 2020}}. \bibinfo{publisher}{Association for Computational
  Linguistics}, \bibinfo{address}{Online}, \bibinfo{pages}{2810--2829}.
\newblock
\urldef\tempurl%
\url{https://doi.org/10.18653/v1/2020.findings-emnlp.253}
\showDOI{\tempurl}


\bibitem[Papineni et~al\mbox{.}(2002)]%
        {BLEU}
\bibfield{author}{\bibinfo{person}{Kishore Papineni}, \bibinfo{person}{Salim
  Roukos}, \bibinfo{person}{Todd Ward}, {and} \bibinfo{person}{Wei{-}Jing
  Zhu}.} \bibinfo{year}{2002}\natexlab{}.
\newblock \showarticletitle{Bleu: a Method for Automatic Evaluation of Machine
  Translation}. In \bibinfo{booktitle}{\emph{Proceedings of the 40th Annual
  Meeting of the Association for Computational Linguistics, July 6-12, 2002,
  Philadelphia, PA, {USA}}}. \bibinfo{publisher}{{ACL}},
  \bibinfo{pages}{311--318}.
\newblock


\bibitem[Park et~al\mbox{.}(2018)]%
        {VQAX}
\bibfield{author}{\bibinfo{person}{Dong~Huk Park}, \bibinfo{person}{Lisa~Anne
  Hendricks}, \bibinfo{person}{Zeynep Akata}, \bibinfo{person}{Anna Rohrbach},
  \bibinfo{person}{Bernt Schiele}, \bibinfo{person}{Trevor Darrell}, {and}
  \bibinfo{person}{Marcus Rohrbach}.} \bibinfo{year}{2018}\natexlab{}.
\newblock \showarticletitle{Multimodal explanations: Justifying decisions and
  pointing to the evidence}. In \bibinfo{booktitle}{\emph{Proceedings of the
  IEEE Conference on Computer Vision and Pattern Recognition}}.
  \bibinfo{pages}{8779--8788}.
\newblock


\bibitem[Patro et~al\mbox{.}(2020)]%
        {robust_explanation}
\bibfield{author}{\bibinfo{person}{Badri Patro}, \bibinfo{person}{Shivansh
  Patel}, {and} \bibinfo{person}{Vinay Namboodiri}.}
  \bibinfo{year}{2020}\natexlab{}.
\newblock \showarticletitle{Robust Explanations for Visual Question Answering}.
  In \bibinfo{booktitle}{\emph{Proceedings of the IEEE/CVF Winter Conference on
  Applications of Computer Vision (WACV)}}.
\newblock


\bibitem[Plummer et~al\mbox{.}(2015)]%
        {flickrEntitie}
\bibfield{author}{\bibinfo{person}{Bryan~A Plummer}, \bibinfo{person}{Liwei
  Wang}, \bibinfo{person}{Chris~M Cervantes}, \bibinfo{person}{Juan~C Caicedo},
  \bibinfo{person}{Julia Hockenmaier}, {and} \bibinfo{person}{Svetlana
  Lazebnik}.} \bibinfo{year}{2015}\natexlab{}.
\newblock \showarticletitle{Flickr30k entities: Collecting region-to-phrase
  correspondences for richer image-to-sentence models}. In
  \bibinfo{booktitle}{\emph{Proceedings of the IEEE international conference on
  computer vision}}. \bibinfo{pages}{2641--2649}.
\newblock


\bibitem[Poth et~al\mbox{.}(2021)]%
        {chunkModel}
\bibfield{author}{\bibinfo{person}{Clifton Poth}, \bibinfo{person}{Jonas
  Pfeiffer}, \bibinfo{person}{Andreas R{\"u}ckl{\'e}}, {and}
  \bibinfo{person}{Iryna Gurevych}.} \bibinfo{year}{2021}\natexlab{}.
\newblock \showarticletitle{What to pre-train on? efficient intermediate task
  selection}.
\newblock \bibinfo{journal}{\emph{arXiv preprint arXiv:2104.08247}}
  (\bibinfo{year}{2021}).
\newblock


\bibitem[Prabhu and Kann(2020)]%
        {disjoint_vocab}
\bibfield{author}{\bibinfo{person}{Nikhil Prabhu} {and}
  \bibinfo{person}{Katharina Kann}.} \bibinfo{year}{2020}\natexlab{}.
\newblock \showarticletitle{Making a Point: Pointer-Generator Transformers for
  Disjoint Vocabularies}. In \bibinfo{booktitle}{\emph{Proceedings of the 1st
  Conference of the Asia-Pacific Chapter of the Association for Computational
  Linguistics and the 10th International Joint Conference on Natural Language
  Processing: Student Research Workshop, {AACL/IJCNLP} 2021, Suzhou, China,
  December 4-7, 2020}}. \bibinfo{pages}{85--92}.
\newblock


\bibitem[Radford et~al\mbox{.}(2019)]%
        {GPT2}
\bibfield{author}{\bibinfo{person}{Alec Radford}, \bibinfo{person}{Jeffrey Wu},
  \bibinfo{person}{Rewon Child}, \bibinfo{person}{David Luan},
  \bibinfo{person}{Dario Amodei}, \bibinfo{person}{Ilya Sutskever},
  {et~al\mbox{.}}} \bibinfo{year}{2019}\natexlab{}.
\newblock \showarticletitle{Language models are unsupervised multitask
  learners}.
\newblock \bibinfo{journal}{\emph{OpenAI blog}} \bibinfo{volume}{1},
  \bibinfo{number}{8} (\bibinfo{year}{2019}), \bibinfo{pages}{9}.
\newblock


\bibitem[Ren et~al\mbox{.}(2015b)]%
        {coco_qa}
\bibfield{author}{\bibinfo{person}{Mengye Ren}, \bibinfo{person}{Ryan Kiros},
  {and} \bibinfo{person}{Richard Zemel}.} \bibinfo{year}{2015}\natexlab{b}.
\newblock \showarticletitle{Exploring models and data for image question
  answering}.
\newblock \bibinfo{journal}{\emph{Advances in neural information processing
  systems}}  \bibinfo{volume}{28} (\bibinfo{year}{2015}).
\newblock


\bibitem[Ren et~al\mbox{.}(2015a)]%
        {fasterRCNN}
\bibfield{author}{\bibinfo{person}{Shaoqing Ren}, \bibinfo{person}{Kaiming He},
  \bibinfo{person}{Ross Girshick}, {and} \bibinfo{person}{Jian Sun}.}
  \bibinfo{year}{2015}\natexlab{a}.
\newblock \showarticletitle{Faster r-cnn: Towards real-time object detection
  with region proposal networks}.
\newblock \bibinfo{journal}{\emph{Advances in neural information processing
  systems}}  \bibinfo{volume}{28} (\bibinfo{year}{2015}).
\newblock


\bibitem[Sammani et~al\mbox{.}(2022)]%
        {GPT_NLE}
\bibfield{author}{\bibinfo{person}{Fawaz Sammani}, \bibinfo{person}{Tanmoy
  Mukherjee}, {and} \bibinfo{person}{Nikos Deligiannis}.}
  \bibinfo{year}{2022}\natexlab{}.
\newblock \showarticletitle{{NLX-GPT:} {A} Model for Natural Language
  Explanations in Vision and Vision-Language Tasks}.
\newblock \bibinfo{journal}{\emph{CoRR}}  \bibinfo{volume}{abs/2203.05081}
  (\bibinfo{year}{2022}).
\newblock


\bibitem[Schuster et~al\mbox{.}(2015)]%
        {scene_graph_textual}
\bibfield{author}{\bibinfo{person}{Sebastian Schuster}, \bibinfo{person}{Ranjay
  Krishna}, \bibinfo{person}{Angel Chang}, \bibinfo{person}{Li Fei-Fei}, {and}
  \bibinfo{person}{Christopher~D Manning}.} \bibinfo{year}{2015}\natexlab{}.
\newblock \showarticletitle{Generating semantically precise scene graphs from
  textual descriptions for improved image retrieval}. In
  \bibinfo{booktitle}{\emph{Proceedings of the fourth workshop on vision and
  language}}. \bibinfo{pages}{70--80}.
\newblock


\bibitem[See et~al\mbox{.}(2017)]%
        {pointer}
\bibfield{author}{\bibinfo{person}{Abigail See}, \bibinfo{person}{Peter~J.
  Liu}, {and} \bibinfo{person}{Christopher~D. Manning}.}
  \bibinfo{year}{2017}\natexlab{}.
\newblock \showarticletitle{Get To The Point: Summarization with
  Pointer-Generator Networks}. In \bibinfo{booktitle}{\emph{Proceedings of the
  55th Annual Meeting of the Association for Computational Linguistics, {ACL}
  2017, Vancouver, Canada, July 30 - August 4, Volume 1: Long Papers}}.
  \bibinfo{pages}{1073--1083}.
\newblock


\bibitem[Selvaraju et~al\mbox{.}(2017)]%
        {grad_cam}
\bibfield{author}{\bibinfo{person}{Ramprasaath~R. Selvaraju},
  \bibinfo{person}{Michael Cogswell}, \bibinfo{person}{Abhishek Das},
  \bibinfo{person}{Ramakrishna Vedantam}, \bibinfo{person}{Devi Parikh}, {and}
  \bibinfo{person}{Dhruv Batra}.} \bibinfo{year}{2017}\natexlab{}.
\newblock \showarticletitle{Grad-CAM: Visual Explanations from Deep Networks
  via Gradient-Based Localization}. In \bibinfo{booktitle}{\emph{{IEEE}
  International Conference on Computer Vision, {ICCV} 2017, Venice, Italy,
  October 22-29, 2017}}. \bibinfo{pages}{618--626}.
\newblock


\bibitem[Tjong Kim~Sang and Buchholz(2000)]%
        {cnll2000}
\bibfield{author}{\bibinfo{person}{Erik~F. Tjong Kim~Sang} {and}
  \bibinfo{person}{Sabine Buchholz}.} \bibinfo{year}{2000}\natexlab{}.
\newblock \showarticletitle{Introduction to the {C}o{NLL}-2000 Shared Task
  Chunking}. In \bibinfo{booktitle}{\emph{Fourth Conference on Computational
  Natural Language Learning and the Second Learning Language in Logic
  Workshop}}.
\newblock


\bibitem[Vaswani et~al\mbox{.}(2017)]%
        {attn_is_all}
\bibfield{author}{\bibinfo{person}{Ashish Vaswani}, \bibinfo{person}{Noam
  Shazeer}, \bibinfo{person}{Niki Parmar}, \bibinfo{person}{Jakob Uszkoreit},
  \bibinfo{person}{Llion Jones}, \bibinfo{person}{Aidan~N Gomez},
  \bibinfo{person}{{\L}ukasz Kaiser}, {and} \bibinfo{person}{Illia
  Polosukhin}.} \bibinfo{year}{2017}\natexlab{}.
\newblock \showarticletitle{Attention is all you need}.
\newblock \bibinfo{journal}{\emph{Advances in neural information processing
  systems}}  \bibinfo{volume}{30} (\bibinfo{year}{2017}).
\newblock


\bibitem[Vedantam et~al\mbox{.}(2015)]%
        {CIDER}
\bibfield{author}{\bibinfo{person}{Ramakrishna Vedantam},
  \bibinfo{person}{C.~Lawrence Zitnick}, {and} \bibinfo{person}{Devi Parikh}.}
  \bibinfo{year}{2015}\natexlab{}.
\newblock \showarticletitle{CIDEr: Consensus-based image description
  evaluation}. In \bibinfo{booktitle}{\emph{{IEEE} Conference on Computer
  Vision and Pattern Recognition, {CVPR} 2015, Boston, MA, USA, June 7-12,
  2015}}. \bibinfo{publisher}{{IEEE} Computer Society},
  \bibinfo{pages}{4566--4575}.
\newblock


\bibitem[Wang et~al\mbox{.}(2022)]%
        {unified_mm_pretrain}
\bibfield{author}{\bibinfo{person}{Peng Wang}, \bibinfo{person}{An Yang},
  \bibinfo{person}{Rui Men}, \bibinfo{person}{Junyang Lin},
  \bibinfo{person}{Shuai Bai}, \bibinfo{person}{Zhikang Li},
  \bibinfo{person}{Jianxin Ma}, \bibinfo{person}{Chang Zhou},
  \bibinfo{person}{Jingren Zhou}, {and} \bibinfo{person}{Hongxia Yang}.}
  \bibinfo{year}{2022}\natexlab{}.
\newblock \showarticletitle{Unifying Architectures, Tasks, and Modalities
  Through a Simple Sequence-to-Sequence Learning Framework}.
\newblock \bibinfo{journal}{\emph{CoRR}}  \bibinfo{volume}{abs/2202.03052}
  (\bibinfo{year}{2022}).
\newblock
\showeprint[arXiv]{2202.03052}


\bibitem[Wang et~al\mbox{.}(2021)]%
        {simvlm}
\bibfield{author}{\bibinfo{person}{Zirui Wang}, \bibinfo{person}{Jiahui Yu},
  \bibinfo{person}{Adams~Wei Yu}, \bibinfo{person}{Zihang Dai},
  \bibinfo{person}{Yulia Tsvetkov}, {and} \bibinfo{person}{Yuan Cao}.}
  \bibinfo{year}{2021}\natexlab{}.
\newblock \showarticletitle{Simvlm: Simple visual language model pretraining
  with weak supervision}.
\newblock \bibinfo{journal}{\emph{arXiv preprint arXiv:2108.10904}}
  (\bibinfo{year}{2021}).
\newblock


\bibitem[Wu and Mooney(2019)]%
        {Faithful_NLE}
\bibfield{author}{\bibinfo{person}{Jialin Wu} {and} \bibinfo{person}{Raymond~J.
  Mooney}.} \bibinfo{year}{2019}\natexlab{}.
\newblock \showarticletitle{Faithful Multimodal Explanation for Visual Question
  Answering}. In \bibinfo{booktitle}{\emph{Proceedings of the 2019 {ACL}
  Workshop BlackboxNLP: Analyzing and Interpreting Neural Networks for NLP,
  BlackboxNLP@ACL 2019, Florence, Italy, August 1, 2019}},
  \bibfield{editor}{\bibinfo{person}{Tal Linzen}, \bibinfo{person}{Grzegorz
  Chrupala}, \bibinfo{person}{Yonatan Belinkov}, {and} \bibinfo{person}{Dieuwke
  Hupkes}} (Eds.). \bibinfo{publisher}{Association for Computational
  Linguistics}, \bibinfo{pages}{103--112}.
\newblock


\bibitem[Xie et~al\mbox{.}(2019)]%
        {vte}
\bibfield{author}{\bibinfo{person}{Ning Xie}, \bibinfo{person}{Farley Lai},
  \bibinfo{person}{Derek Doran}, {and} \bibinfo{person}{Asim Kadav}.}
  \bibinfo{year}{2019}\natexlab{}.
\newblock \showarticletitle{Visual Entailment: {A} Novel Task for Fine-Grained
  Image Understanding}.
\newblock \bibinfo{journal}{\emph{CoRR}}  \bibinfo{volume}{abs/1901.06706}
  (\bibinfo{year}{2019}).
\newblock


\bibitem[Yang et~al\mbox{.}(2018)]%
        {SGP0}
\bibfield{author}{\bibinfo{person}{Jianwei Yang}, \bibinfo{person}{Jiasen Lu},
  \bibinfo{person}{Stefan Lee}, \bibinfo{person}{Dhruv Batra}, {and}
  \bibinfo{person}{Devi Parikh}.} \bibinfo{year}{2018}\natexlab{}.
\newblock \showarticletitle{Graph {R-CNN} for Scene Graph Generation}. In
  \bibinfo{booktitle}{\emph{Computer Vision - {ECCV} 2018 - 15th European
  Conference, Munich, Germany, September 8-14, 2018, Proceedings, Part {I}}}
  \emph{(\bibinfo{series}{Lecture Notes in Computer Science},
  Vol.~\bibinfo{volume}{11205})}, \bibfield{editor}{\bibinfo{person}{Vittorio
  Ferrari}, \bibinfo{person}{Martial Hebert}, \bibinfo{person}{Cristian
  Sminchisescu}, {and} \bibinfo{person}{Yair Weiss}} (Eds.).
  \bibinfo{publisher}{Springer}, \bibinfo{pages}{690--706}.
\newblock


\bibitem[Young et~al\mbox{.}(2014)]%
        {flickr30k}
\bibfield{author}{\bibinfo{person}{Peter Young}, \bibinfo{person}{Alice Lai},
  \bibinfo{person}{Micah Hodosh}, {and} \bibinfo{person}{Julia Hockenmaier}.}
  \bibinfo{year}{2014}\natexlab{}.
\newblock \showarticletitle{From image descriptions to visual denotations: New
  similarity metrics for semantic inference over event descriptions}.
\newblock \bibinfo{journal}{\emph{Transactions of the Association for
  Computational Linguistics}}  \bibinfo{volume}{2} (\bibinfo{year}{2014}),
  \bibinfo{pages}{67--78}.
\newblock


\bibitem[Zellers et~al\mbox{.}(2019)]%
        {VCR}
\bibfield{author}{\bibinfo{person}{Rowan Zellers}, \bibinfo{person}{Yonatan
  Bisk}, \bibinfo{person}{Ali Farhadi}, {and} \bibinfo{person}{Yejin Choi}.}
  \bibinfo{year}{2019}\natexlab{}.
\newblock \showarticletitle{From recognition to cognition: Visual commonsense
  reasoning}. In \bibinfo{booktitle}{\emph{Proceedings of the IEEE/CVF
  conference on computer vision and pattern recognition}}.
  \bibinfo{pages}{6720--6731}.
\newblock


\bibitem[Zhang et~al\mbox{.}(2019)]%
        {SGP1}
\bibfield{author}{\bibinfo{person}{Ji Zhang}, \bibinfo{person}{Kevin~J Shih},
  \bibinfo{person}{Ahmed Elgammal}, \bibinfo{person}{Andrew Tao}, {and}
  \bibinfo{person}{Bryan Catanzaro}.} \bibinfo{year}{2019}\natexlab{}.
\newblock \showarticletitle{Graphical contrastive losses for scene graph
  parsing}. In \bibinfo{booktitle}{\emph{Proceedings of the IEEE/CVF Conference
  on Computer Vision and Pattern Recognition}}. \bibinfo{pages}{11535--11543}.
\newblock


\bibitem[Zhu et~al\mbox{.}(2016)]%
        {visual7w}
\bibfield{author}{\bibinfo{person}{Yuke Zhu}, \bibinfo{person}{Oliver Groth},
  \bibinfo{person}{Michael Bernstein}, {and} \bibinfo{person}{Li Fei-Fei}.}
  \bibinfo{year}{2016}\natexlab{}.
\newblock \showarticletitle{Visual7w: Grounded question answering in images}.
  In \bibinfo{booktitle}{\emph{Proceedings of the IEEE conference on computer
  vision and pattern recognition}}. \bibinfo{pages}{4995--5004}.
\newblock


\end{thebibliography}


\clearpage
\appendix
\section{Failure cases}

\begin{figure*}[]
  \centering
  \includegraphics[width=\linewidth]{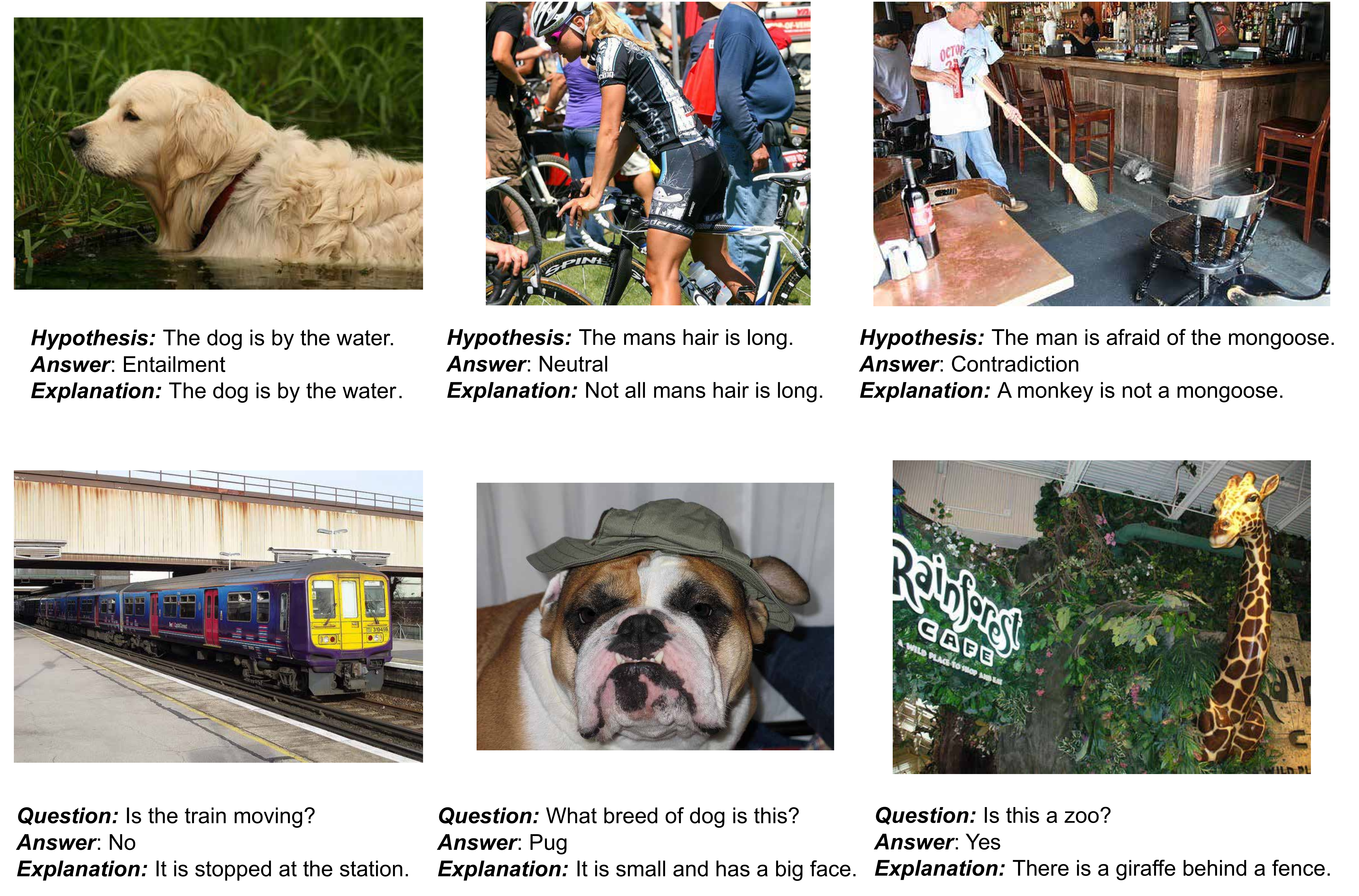}
  \caption{Failure cases on e-SNLI-VE and VQA-X.}
  \label{fig:failuer_cases}
\end{figure*}

We include failure cases on e-SNLI-VE and VQA-X of our model in Figure~\ref{fig:failuer_cases}. 
We observe that the failure cases mainly involve misinterpretation of image details (orientation between objects, the gender of the people, the breed of the animals, and the characters).
These cases show that although CALeC can exploit the rich semantics contained in phrase thorough chunk-aware semantic interactor, it still has limitations on the image comprehension, which can be a future direction of our work.
For e-SNLI-VE, we observe that if the relationship is \textit{entailment}, the model tends to repeat the hypothesis, which may result from the bias of the dataset.
Although the answers are predicted wrong, the explanations are faithful to the answers, which shows that the lexical constraint-aware generator can reflect the decision-making process and help correct the model bias.

\end{document}